\documentclass{article} 
\usepackage[preprint]{neurips_2025}

\usepackage{amsmath,amsfonts,bm}

\def\eqref#1{equation~\ref{#1}}

\def\1{\bm{1}}

\DeclareMathAlphabet{\mathsfit}{\encodingdefault}{\sfdefault}{m}{sl}
\SetMathAlphabet{\mathsfit}{bold}{\encodingdefault}{\sfdefault}{bx}{n}

\usepackage{enumitem}
\usepackage[hidelinks]{hyperref}
\usepackage{url}
\usepackage{booktabs}
\usepackage{amsfonts}
\usepackage{graphicx}
\usepackage{sidecap}
\usepackage[small]{caption}
\usepackage{subcaption}
\usepackage{amsmath}
\allowdisplaybreaks
\usepackage{amsthm}

\usepackage{wrapfig}
\usepackage{booktabs}
\usepackage{multicol}

\usepackage[compact]{titlesec}
\titlespacing{\section}{0pt}{1ex}{0.5ex}
\titlespacing{\subsection}{0pt}{0.5ex}{0ex}
\titlespacing{\subsubsection}{0pt}{0.5ex}{0ex} 
\usepackage{setspace}
\AtBeginDocument{%
  \addtolength\abovedisplayskip{-0.25\baselineskip}%
  \addtolength\belowdisplayskip{-0.25\baselineskip}%
  \addtolength\abovedisplayshortskip{-0.25\baselineskip}%
  \addtolength\belowdisplayshortskip{-0.25\baselineskip}%
}

\usepackage{pifont}
\usepackage{xspace}
\newcommand{\circone}{\ding{172}\xspace}
\newcommand{\circtwo}{\ding{173}\xspace}
\newcommand{\circthree}{\ding{174}\xspace}
\newcommand{\circfour}{\ding{175}\xspace}

\usepackage{tabularx, booktabs}
\newcolumntype{C}{>{\centering\arraybackslash}X}
\newcolumntype{R}{>{\raggedleft\arraybackslash}X}
\newcolumntype{S}{>{\raggedleft\arraybackslash\hsize=.5\hsize}X}

\usepackage{arydshln}
\usepackage{booktabs}
\usepackage{multirow}

\usepackage{pifont}  %
\usepackage{makecell}

\newcommand{\optparens}[1]{\if\relax\detokenize{#1}\relax\else(#1)\fi}   %

\usepackage[noabbrev,capitalize]{cleveref}
\crefname{equation}{equation}{equations}
\crefname{section}{section}{sections}
\crefname{footnote}{footnote}{footnotes}   
\crefname{line}{line}{lines}   
\crefname{assumption}{assumption}{assumptions}
\crefname{lstlisting}{listing}{listings}
\Crefname{lstlisting}{Listing}{Listings}
\usepackage{bbm}
\usepackage{appendix}

\usepackage{xcolor}  %
\usepackage{soul}
\definecolor{aigold}{RGB}{244,210, 1} 
\definecolor{aigreen}{RGB}{245, 255, 249}

\definecolor{humanpurple}{RGB}{235, 222, 240} 

\definecolor{commentgray}{RGB}{86, 101, 115}

\definecolor{light-blue}{rgb}{0.6,0.6,1}

\definecolor{aired}{RGB}{255,180,181}

\usepackage{listings}
\usepackage{parcolumns}
\lstdefinestyle{datalogstyle}{
	basicstyle={\codefont\small},  
	xleftmargin={6pt},
        xrightmargin={6pt},
        breakindent=0pt,
	frame=tb,
	stepnumber=1,
	firstnumber=1,
	numberfirstline=true,
	tabsize=2,
	showtabs=false,
	showspaces=false,
	showstringspaces=false,
	extendedchars=true,
	breaklines=true,
	columns=fullflexible,
	keepspaces=true,
	escapeinside={@}{@},
	firstnumber=last,
	captionpos=b,
	commentstyle=\color{black!65},
	numberstyle=\tiny\color{black!65},
	stringstyle=\color{codepurple},
	breakatwhitespace=false, 
	keepspaces=true,                 
	numbersep=5pt,                  
	showspaces=false,                
	showstringspaces=false,
	showtabs=false,
	aboveskip={0.8\baselineskip},
	belowskip={0.2\baselineskip},
	backgroundcolor=\color{aigreen},
}
\definecolor{rebuttal}{RGB}{229,255,204}
\sethlcolor{rebuttal}

\newcommand{\codefont}{\fontfamily{lmtt}\selectfont}

\usepackage[textwidth=2.7cm]{todonotes}

\newcommand{\cutforspace}[1]{}

\newcommand\myshade{85}
\colorlet{myurlcolor}{blue}
\hypersetup{
  citecolor  = black,
  urlcolor   = myurlcolor!\myshade!black,
  colorlinks = true,
}

\newcommand{\datatitle}[1]{{#1}}

\title{FAMMA: A Benchmark for \underline{F}in\underline{a}ncial \underline{M}ultilingual \underline{M}ultimodal Question \underline{A}nswering}

\author{Siqiao Xue$^{\diamondsuit}$, Xiaojing  Li$^{\diamondsuit}$, Fan Zhou$^{\diamondsuit}$, Qingyang Dai$^{\diamondsuit}$, Zhixuan Chu$^{\spadesuit}$, Hongyuan Mei$^{\clubsuit\heartsuit}$ \\
$^{\diamondsuit}$Alipay \quad $^{\spadesuit}$Zhejiang University \quad $^{\clubsuit}$TTIC \quad $^{\heartsuit}$Purdue University\\
\texttt{\{siqiao.xsq,sommerlxj,moutozf,jerrymccree520\}@gmail.com}\\
\texttt{zhixuanchu@zju.edu.cn, hongyuan@ttic.edu} }

\begin{document}
\maketitle

\begin{abstract}
In this paper, we introduce FAMMA, an open-source benchmark for \underline{f}in\underline{a}ncial \underline{m}ultilingual \underline{m}ultimodal question \underline{a}nswering (QA).\footnote{This name is made to honor Eugene Fama, ``the father of modern finance'' and a Nobel prize winner.}
Our benchmark aims to evaluate the abilities of large language models (LLMs) in answering complex reasoning questions that require advanced financial knowledge. 
The benchmark has two versions: FAMMA-Basic consists of 1,945 questions extracted from university textbooks and exams, along with human-annotated answers and rationales; FAMMA-LivePro consists of 103 novel questions created by human domain experts, with answers and rationales held out from the public for a contamination-free evaluation.
These questions cover advanced knowledge of 8 major subfields in finance (e.g., corporate finance, derivatives, and portfolio management).
Some are in Chinese or French, while a majority of them are in English.
Each question has some non-text data such as charts, diagrams, or tables.
Our experiments reveal that FAMMA poses a significant challenge on LLMs, including reasoning models such as GPT-o1 and DeepSeek-R1.
Additionally, we curated 1,270 reasoning trajectories of DeepSeek-R1 on the FAMMA-Basic data, and fine-tuned a series of open-source Qwen models using this reasoning data.
We found that training a model on these reasoning trajectories can significantly improve its performance on FAMMA-LivePro.
We released our leaderboard, data, code, and trained models at {\small\url{https://famma-bench.github.io/famma/}}.
\end{abstract}

\section{Introduction}
\label{sec:intro}
Benchmarks have played a pivotal role in advancing AI research, especially in the realm of large language models (LLMs)~\citep{brown-2020-gpt,gpt4,touvron2023llama2,jiang2023mistral,jiang2024mixtral,meta2024llama}.
Benchmarks have been helping researchers track the advancement of LLMs in a variety of capabilities, including 
general language understanding and knowledge acquisition~\citep{wang2019glue,hendrycks2020measuring,zhou2023instruction,wang2024mmlu}, 
code generation~\citep{chen2021evaluating,evalplus,jimenez2023swe},
mathematical reasoning~\citep{cobbe2021training,hendrycksmath2021}, 
tool use~\citep{bfcl,srinivasan2023nexusraven,trivedi2024appworld}, 
and legal reasoning~\citep{guha2023legalbench}. 
However, we have seen a scarcity of high-quality benchmarks in financial reasoning, an area where both researchers and practitioners will benefit from LLMs.

\begin{figure}[t]
    \centering
    \subfloat[A medium-level question in financial statement analysis, requiring knowledge of accounting and corporate finance. 
    Often seen in CFA-Level II exams.
    ]{%
        \includegraphics[width=0.48\textwidth]{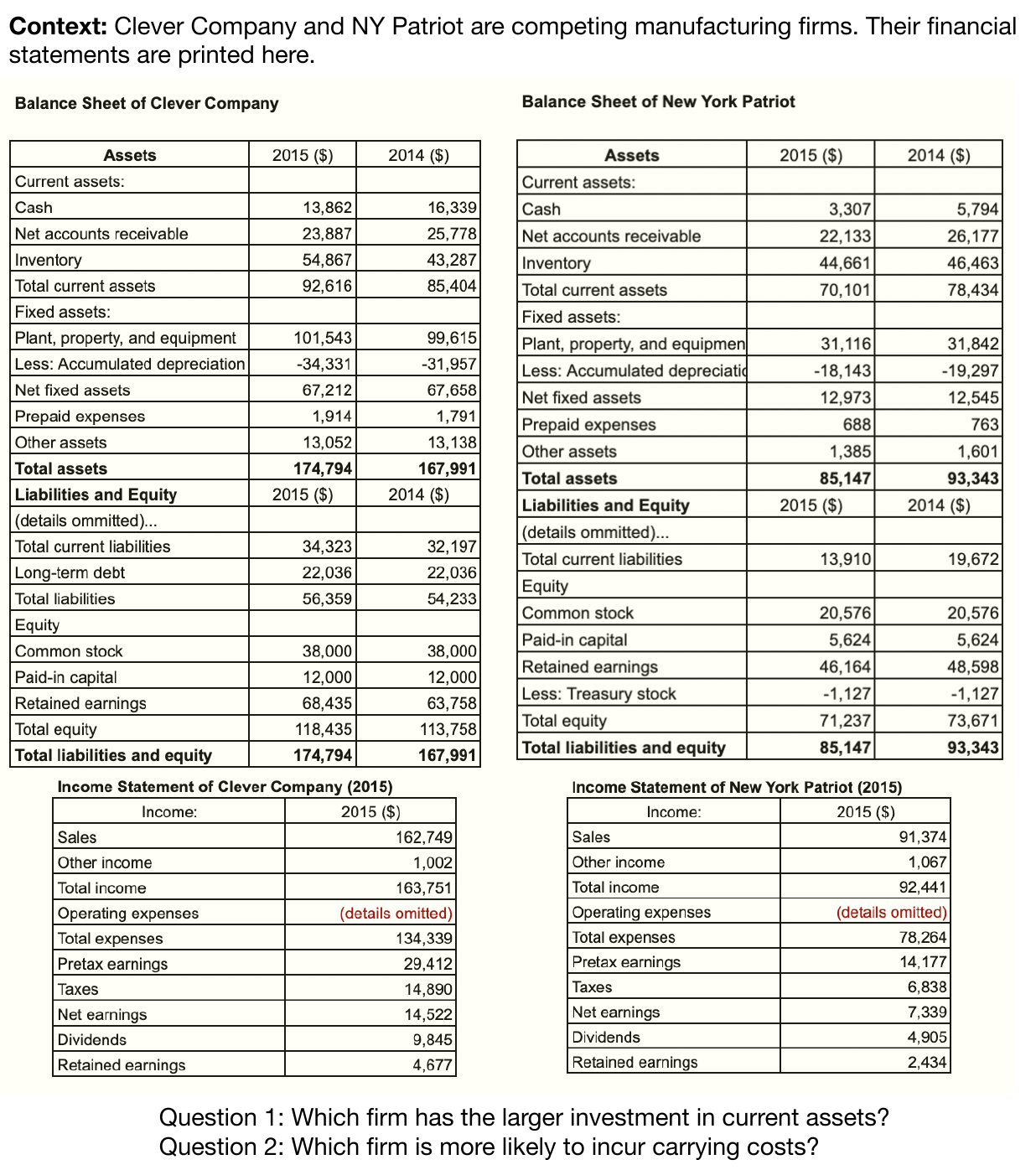}%
        \label{fig:example_1}%
    }
    \hfill
    \subfloat[A hard-level question in derivatives, requiring knowledge of probability theory and stochastic calculus. Often seen in CFA-Level III exams.
    ]
    {%
        \includegraphics[width=0.51\textwidth]{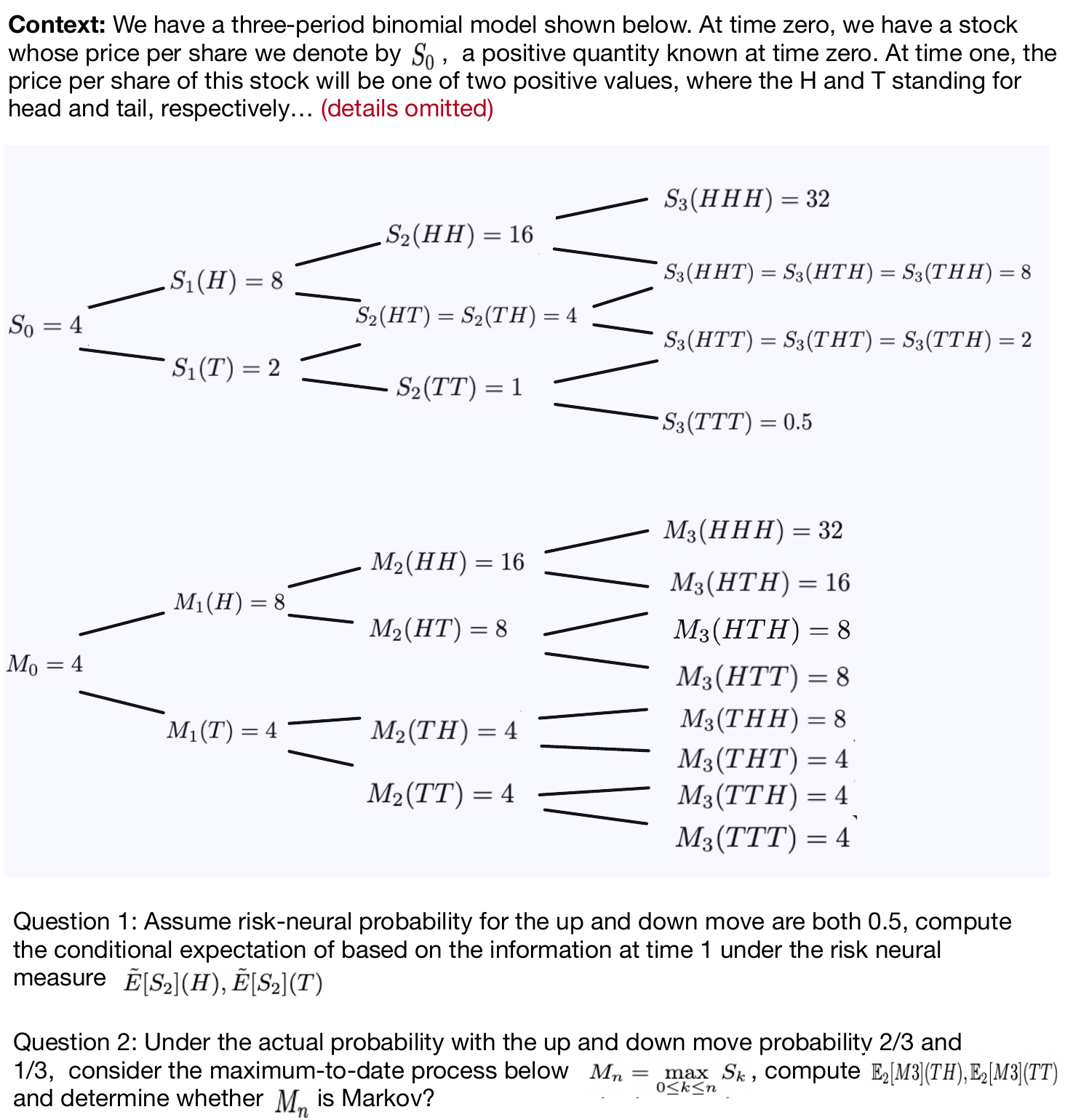}%
        \label{fig:example_2}%
    }
    \caption{Questions in FAMMA, requiring highly specialized knowledge and sophisticated calculation.}
    \label{fig:bench_example}
\end{figure}

Real-world financial reasoning problems are uniquely interesting because they often require both highly specialized knowledge and sophisticated calculation; see examples in \cref{fig:bench_example}.
This characteristic distinguishes such problems from pure arithmetic reasoning problems (e.g., GSM8K~\citep{cobbe2021training} and MATH~\citep{lightman2023let}) and pure knowledge-heavy problems (e.g., GPQA~\citep{rein2024gpqa}).
Therefore, we constructed \datatitle{FAMMA}, an open-source benchmark for \underline{f}in\underline{a}ncial \underline{m}ultilingual \underline{m}ultimodal question \underline{a}nswering.
FAMMA consists of real-world problems that financial practitioners need to solve in their daily work: e.g., financial analysis problems that require understanding complicated balance sheets, option pricing problems that require advanced stochastic calculus knowledge.
The FAMMA-Basic version includes 1,975 meticulously collected from university textbooks and exams, along with human-annotated answers and rationales; the FAMMA-LivePro version consists of 103 novel questions created by human experts, with answers and rationales held out from the public for a contamination-free evaluation.
The questions cover advanced knowledge of 8 major subfields in finance, including corporate finance, asset management, and financial engineering.
Though most questions are in English, some of them are in Chinese and French.
Each question has some non-text data (e.g., charts, diagrams, and tables).




To facilitate further research, we released text-only variants of both FAMMA-Basic and FAMMA-LivePro. These versions employ OCR to extract multimodal content--such as text embedded in charts, formulas, and figures--and convert it into plain textual context. This transformation enables the use of text-only models and simplifies integration with language-model-based pipelines.

In addition to the benchmark, we curated a set of reasoning trajectories that correctly solve 1270 FAMMA-Basic questions through prompting DeepSeek-R1.
We fine-tuned a series of open-source Qwen models using this reasoning data, and found that training a model on these reasoning trajectories can significantly improve its performance on FAMMA-LivePro. A full summary of FAMMA versions, including reasoning and OCR-based variants, is provided in \cref{tab:famma_version}.

We evaluated a range of LLMs on FAMMA, including strong reasoning models such as GPT-o1 and DeepSeek-R1. 
Our key findings include:
\circone FAMMA poses a significant challenge for strong models: e.g., GPT-o1 can only solve 30\% of the hard questions of FAMMA-LivePro;
\circtwo it will significantly improve the accuracy of a model on the arithmetic questions if we prompt it to generate a Python program for numerical calculation and then execute the program;
\circthree all the models that we tested struggle at solving knowledge-heavy non-arithmetic questions;
\circfour retrieval-augmented-generation doesn't provide additional gains to reasoning models like GPT-o1 and DeepSeek-R1.


\begin{table*}[ht]
	\begin{center}
		\begin{small}
			\begin{sc}
				\begin{tabularx}{\textwidth}{l C C C C C}
					\toprule
				Publish Date	  & Version &  Size  & \multicolumn{3}{c}{Source}   \\
					\midrule
					2024-10-06 & Basic & $1945$ & \multicolumn{3}{c}{\scriptsize  Public online sources.} \\
                    2025-02-01 & LivePro & $103$ & \multicolumn{3}{c}{\scriptsize  Authored by domain experts.} \\
                    2025-03-01 & Basic-Txt & $1945$ & \multicolumn{3}{c}{\scriptsize OCR-converted textual version of \texttt{Basic}.} \\
                    2025-03-01 & LivePro-Txt & $103$ & \multicolumn{3}{c}{\scriptsize OCR-converted textual version of \texttt{LivePro}.} \\
                    2025-04-22 & Reasoning & $1273$ & \multicolumn{3}{c}{\scriptsize Reasoning traces distilled from DeepSeek-R1.} \\
					\bottomrule
				\end{tabularx}
			\end{sc}
		\end{small}
	\end{center}
	\caption{Overview of FAMMA Versions.}
	\label{tab:famma_version}
\end{table*}

\section{Related Work}
\label{sec:related_work}
The application of natural language technologies in finance dates back to the early 2000s, when sentiment analysis was used to analyze how media would impact stock market movements~\citep{tetlock2007giving,pang2008opinion}. 
Over recent years, the emergence of LLMs has inspired research in advancing financial industry with LLMs, 
including pretraining and fine-tuning LLMs with finance-related text~\citep{wu2023bloomberggptlargelanguagemodel,yang2023fingptopensourcefinanciallarge}, 
improving sentiment analysis with LLMs~\citep{finllama2024,inserte2024largelanguagemodeladaptation,cao2024risklabspredictingfinancialrisk}, 
and building chatbots that specialize in finance knowledge~\citep{langchain,finchat,xue2023weaverbird,xue2024demonstration}. 

Several existing benchmarks can be used to evaluate these modern models and systems, including 
FiQA~\citep{fiqa2018}, 
FinQA~\citep{chen2021finqa}, 
ConvFinQA~\citep{chen2022convfinqa}, 
FinanceBench~\citep{islam2023financebench}, 
and FinBen~\citep{xie2024finben}. 
However, these benchmarks are not as challenging as real-world financial reasoning problems. 
In particular, they only have text-based questions; they are all in English; answering their questions only requires knowledge at a rudimentary to intermediate level~\citep{hendrycks2020measuring,chen2021finqa,islam2023financebench,xie2024finben}.
The finance-related questions in MMMU~\citep{yue2023mmmu} involve data of other modalities like tables and charts. But this general-purpose benchmark covers multiple disciplines (e.g., art, business, science, humanities, etc), and thus has a very limited coverage on finance-related questions. 
Our FAMMA benchmark makes a unique and focused contribution to the community on top of existing benchmarks: it has a much broader coverage on subfields of finance; its data is in multiple languages and of multiple modalities; its questions test advanced knowledge.


\section{The FAMMA Benchmark}\label{sec:famma}

FAMMA provides comprehensive coverage across 8 key subfields: alternative investments, corporate finance, derivatives, economics, equity, financial statement analysis, fixed income, and portfolio management. These topics closely align with those taught in elite academic programs (e.g., Princeton's \href{https://bcf.princeton.edu/academic-programs/master-in-finance/new-students/requirements-and-core-courses/}{\textcolor{blue}{Masters in Finance}} program) as well as professional certifications (e.g., \href{https://www.cfainstitute.org/}{\textcolor{blue}{CFA}}).
The dataset consists of both multiple-choice ($52.3\%$) and open questions ($47.7\%$). Additionally, $70.4\%$ of the questions feature single-image scenarios, while $29.6\%$ involve multi-image scenarios. Notably, $99.5\%$ of the questions are accompanied by explanations. The questions are distributed across three difficulty levels and three most widely used languages in the finance industry~\citep{efinancialcareers2022}: English ($68.0\%$), Chinese ($22.8\%$), and French ($9.2\%$).
The overall subject coverage and statistics are shown in \cref{tab:famma_stat}. 



\begin{table*}[ht]
	\begin{center}
		\begin{small}
			\begin{sc}
				\begin{tabularx}{\textwidth}{l *{2}{C}*{2}{C}}
					\toprule
					 & \multicolumn{2}{c}{\# Questions} & \multicolumn{2}{c}{\# Tokens Per Question} \\
					\cmidrule(lr){2-3} \cmidrule(lr){4-5}
				Breakdown	  & FAMMA-Basic &  FAMMA-LivePro  & FAMMA-Basic &  FAMMA-LivePro   \\
					\midrule
					Overall & $1945$ & $103$ & $247.9$ & $125.1$ \\
                    By types &   &   \\
                    \quad \scriptsize Multiple-choice & \scriptsize $1057$ & \scriptsize $15$  & \scriptsize $368.3$ & \scriptsize $70.3$ \\
                    \quad \scriptsize Open-ended & \scriptsize $888$ & \scriptsize $88$ & \scriptsize $104.7$& \scriptsize $130.0$  \\
                    By arithmeticity &    & \\
                    \quad \scriptsize arithmetic & \scriptsize $1075$ & \scriptsize $46$ & \scriptsize $383.1$  &\scriptsize $106.6$ \\
                    \quad \scriptsize non-arithmetic & \scriptsize $870$ & \scriptsize $57$ & \scriptsize $138.5$ & \scriptsize $135.6$ \\
                    By difficulties &  &  \\
                    \quad \scriptsize Easy &    \scriptsize $694$ & \scriptsize $34$ & \scriptsize $135.6$ & \scriptsize $100.4$  \\
                    \quad \scriptsize Medium & \scriptsize $473$ & \scriptsize $39$  &  \scriptsize $182.5$ & \scriptsize $165.8$ \\
                    \quad \scriptsize Hard & \scriptsize $778$ & \scriptsize $30$ & \scriptsize $387.9$ & \scriptsize $106.3$  \\

                    By languages &  &    \\
                    \quad \scriptsize English & \scriptsize $1378$ & \scriptsize $35$  & \scriptsize $256.3$ &  \scriptsize $116.3$  \\
                    \quad \scriptsize Chinese & \scriptsize $441$ & \scriptsize $34$ & \scriptsize $217.3$& \scriptsize $130.4$   \\
                    \quad \scriptsize French & \scriptsize $156$ & \scriptsize $34$  & \scriptsize $254.9$&  \scriptsize $137.1$   \\
                    By subfields &  &    \\
                    \quad \scriptsize Alternative Investments & \scriptsize $92$  & \scriptsize n.a.  & \scriptsize $483.4$ & \scriptsize n.a.   \\
                    \quad \scriptsize Corporate Finance & \scriptsize $269$   & \scriptsize n.a. & \scriptsize $84.4$ & \scriptsize n.a.   \\
                    \quad \scriptsize Derivatives & \scriptsize $296$  & \scriptsize $36$ & \scriptsize $292.8$  & \scriptsize $110.0$   \\
                    \quad \scriptsize Economics & \scriptsize $44$      & \scriptsize $15$ & \scriptsize $601.8$ & \scriptsize $310.5$  \\
                    \quad \scriptsize Equity & \scriptsize $278$ &\scriptsize n.a.  &\scriptsize $256.5$ & \scriptsize n.a.    \\
                    \quad \scriptsize Financial Statement Analysis & \scriptsize $100$   & \scriptsize n.a.   & \scriptsize $52.3$ & \scriptsize n.a. \\
                    \quad \scriptsize Fixed Income &   \scriptsize $264$  & \scriptsize $34$  &\scriptsize $224.8$ & \scriptsize $73.9$   \\
                    \quad \scriptsize Portfolio Management & \scriptsize $602$  & \scriptsize $18$    & \scriptsize $275.7$ & \scriptsize $119.2$  \\
					\bottomrule
				\end{tabularx}
			\end{sc}
		\end{small}
	\end{center}
	\caption{Key statistics of FAMMA.}
	\label{tab:famma_stat}
\end{table*}

\subsection{Benchmark Construction}\label{sec:benchmark}

\paragraph{FAMMA-Basic collection.} We assemble a team of seven volunteer STEM researchers to create a comprehensive set of multimodal questions. Five are co-authors of this paper, while the other two are graduates from a Chinese university. Two annotators hold finance degrees, and the others have completed relevant coursework. Two annotators are trilingual (English, Chinese, and French), while the rest are bilingual.
These annotators draw upon open-source textbooks, exams, and other study materials (see \cref{tab:data_source} in \cref{app:dataset} for details), and apply their expertise to rewrite or create new questions when needed. The new questions are either entirely original, not present in the data sources, or enhanced versions of existing questions. 




The annotators are tasked with selecting questions that require advanced, master-level, or professional knowledge to answer. This selection process is guided by aligning the questions with a minimum of CFA Level 1 difficulty~\citep{cfal1}, ensuring they meet industry standards of complexity. Additionally, selected questions must incorporate multimodal information, such as tables, images, or other visual data, to enrich the input and challenge the model's ability to process diverse formats. By following these criteria, we have curated a diverse set of approximately 2,100 questions, drawn from a wide range of authoritative sources.

\paragraph{Data quality control on FAMMA-Basic.}
To ensure high data quality we apply a two-stage cleaning protocol, carried out by three annotators: two who independently complete the annotations and a third who subsequently validates their work. 

\begin{itemize}[leftmargin=*]
\item In the first stage, all 2,100 candidate questions are manually inspected to correct formatting errors, fix typos, remove duplicates, and verify explanation accuracy. During this process: 
\begin{itemize}[leftmargin=*]
    \item $8\%$ of the questions were flagged for formatting issues, including UTF encoding errors (notably in Chinese and French) and improper rendering of mathematical expressions. These issues were corrected and the questions retained for the second stage.
    \item $4\%$ of the questions were removed due to duplication or irrelevance.
    \item $13\%$ of the explanations were revised or fully rewritten to improve clarity, correctness, or alignment with the question content.
\end{itemize}

\item Each surviving item receives four semantic tags: question type (multiple-choice / open), arithmeticity (requires numeric calculation or not), language, and --- our main focus--\emph{difficulty tier} and \emph{subfield}.  
      The first three labels are largely mechanical and agreed upon almost instantly; we therefore detail only the latter two:


\begin{itemize}[leftmargin=*]
    \item The difficulty levels are aligned with the concept-specific standards of the CFA curriculum~\citep{cfaguideline}. In addition, questions demanding richer information processing (e.g., multiple tables or figures) are likewise placed in higher difficulties. Two primary annotators labelled the entire set independently, while a third served as arbitrator when disagreements arose. The two primaries agreed outright on roughly $80\%$ of the items (inter-annotator consistency is high, with Cohen’s $\kappa = 0.78$.); residual conflicts were resolved by discussion. Questions deemed overly simplistic--- purely on memorization or with answers that are obvious from the context---are removed ($1\%$) to preserve challenge. 
    \item The subfield annotation is determined by the explicit topics provided in the data source. If not clearly specified, the primary annotators use their discretion to assign the most appropriate category based on the content of the question, and the third annotator validates the choice.
\end{itemize}
\end{itemize}

The final corpus contains 1,945 carefully curated questions.
Summary statistics---broken down by question type, arithmeticity, difficulty, language, and subfield---are reported in \cref{tab:famma_stat}.
JSON schemas are provided in \Cref{lst:mc_question_json_format} for multiple-choice items and in \Cref{lst:open_question_json_format} for open-ended items; complete examples are included in \cref{app:dataset}.


\paragraph{FAMMA-LivePro collection.}
FAMMA–LivePro extends the benchmark with a set of entirely new questions authored by four multilingual senior practitioners in quantitative finance.  
Each expert has 5–10 years of experience at leading firms and holds a master’s degree or Ph.D.; detailed biographies appear in \cref{app:dataset}. The creation pipeline mirrors that of FAMMA–Basic.
Three practitioners first drafted the questions in English, generating seven core problem templates that were expanded into full questions---none drawn from existing exams or public materials.
These English sources then passed through the two-stage cleaning protocol: the two primary annotators performed formatting fixes and initial labelling, while the fourth expert served as validator.
Once finalised, the same team translated every question into Chinese and French and reapplied the identical annotation steps, guaranteeing uniform metadata and quality across all three languages.

The result is a contamination-free release of 103 questions; all ground-truth answers and rationales are withheld from public distribution to preserve the integrity of future evaluations.



\paragraph{Text-only FAMMA versions collection.} Using PaddleOCR \citep{paddleocr}, we convert multimodal elements (image, tables, etc) in the original datasets to plain text, yielding two new variants: FAMMA-Basic-Txt and FAMMA-LivePro-Txt. These versions enable straightforward evaluation of open-sourced reasoning models and their distilled variants that accept text input only.

\subsection{Collection of reasoning trajectories}\label{sec:reasoningdata}

The reasoning trajectories are generated by a dedicated distillation pipeline:

\begin{itemize}[leftmargin=*]
    \item Base corpus. We use FAMMA-Basic-Txt, with all images and tables OCR-converted into text.
    
    \item Distillation. Each question is fed to DeepSeek-R1 together with a prompt that elicits two parallel outputs:  
    (i) a free-form, long-form chain-of-thought (CoT) reasoning trace, and  
    (ii) a formally tagged execution trace that interleaves $\langle$think$\rangle$, $\langle$search$\rangle$, $\langle$python$\rangle$, and $\langle$information$\rangle$ actions.

    \item Tool execution. Whenever the model outputs a $\langle$search$\rangle$ or $\langle$python$\rangle$ tag, the associated query or code is executed within a sandbox environment.  
    The returned snippet or stdout is fed back to the agent so that the model can continue reasoning. The full interaction is recorded as the ``structured thinking trajectory,'' while the cleaned narrative forms the ``thinking trajectory.''
    
    \item Verification. After  $\langle$answer$\rangle$ tag is issued, the final answer is automatically verified by GPT-4o against the ground-truth labels in FAMMA.  
    Only correctly solved runs are retained, ensuring logically coherent, self-consistent reasoning paths.

\end{itemize}

This pipeline yields a corpus of 1,273 multilingual CoT traces, revealing both the final answer and the complete reasoning path.


\section{Experiments}
\subsection{Experimental Setup}
\label{sec:exp_setup}

\paragraph{Benchmarked MLLMs.} We evaluate a diverse suite of state-of-the-art models on both FAMMA–Basic and FAMMA–LivePro:

\begin{itemize}[leftmargin=*]
    \item Proprietary models.  
    GPT-o1~\citep{gpt4o1}, GPT-4o~\citep{gpt4o}, Claude-3.5-Sonnet and Claude-3-Sonnet~\citep{claudetechreport2024,claudetechreport2023},  
          Gemini-2.0-Pro, Gemini-2.0-Flash-Thinking, and Gemini-1.5-Pro~\citep{gemini_2,gemini_thinking,gemini}.  
          All seven are the top-ranked on the LMSYS Multimodal-Arena leaderboard~\citep{lmsys}. 

    \item Open-source models.  We benchmark two publicly released model lines:  
    \begin{itemize}[leftmargin=*]
        \item Qwen2.5-VL (7 B–72 B parameters), which holds state-of-the-art scores on vision–language tasks such as MathVista~\citep{lu2024mathvista} and DocVQA~\citep{docvqa2021}.  
      \item DeepSeek-R1 (and its distilled Qwen variants) together with Qwen-QwQ-32B.  Both are the top-ranked open-weight entries on the LMSYS Arena leaderboard and rival GPT-o1 on reasoning-focused evaluations including AIME-2024 and MATH-500.  
              Because these models accept text only, we first convert all images and tables to plain text with PaddleOCR \citep{paddleocr}. Models evaluated with this OCR preprocessing are marked with an asterisk (*).
    \end{itemize}
    
\end{itemize}

Additionally, we apply Program-of-Thoughts (PoT) prompting \citep{chen2022program} to selected top models , guiding them to generate and execute Python code for arithmetic questions. Models that use this technique are prefixed with ``PoT''.

\paragraph{Generation process.} MLMMs are instructed to understand the format and the structure of the questions, and return the response, under a zero-shot setting on our benchmark. The instruction prompts are designed to be straightforward and consistent across all models.
Please refer to \Cref{lst:mc_instruction_format} and \Cref{lst:open_instruction_format} in \cref{app:exp} for the prompts used to to guide responses to multiple-choice and open questions, respectively. During the final stage of generation for multiple-choice questions, we utilize both regex and GPT-4o to extract the corresponding lettered option from the response. Any discrepancies between the two methods will be manually reviewed and validated by annotators.

\paragraph{LM-powered evaluation.}
During the evaluation process, we use GPT-4o as an LM evaluator to assess the accuracy of responses generated by LLMs for each question. The reported score represents the accuracy of these responses. Each response is categorized as either correct or incorrect, and the reported score reflects the average accuracy across the entire set of questions. We set the temperature of the LM evaluator as $0$ to keep the evaluation results deterministic. See \cref{app:llm_eval} for more details.

\begin{figure}[htbp]
    \centering
    \begin{minipage}[t]{0.49\textwidth}
        \centering
        \includegraphics[width=\textwidth]{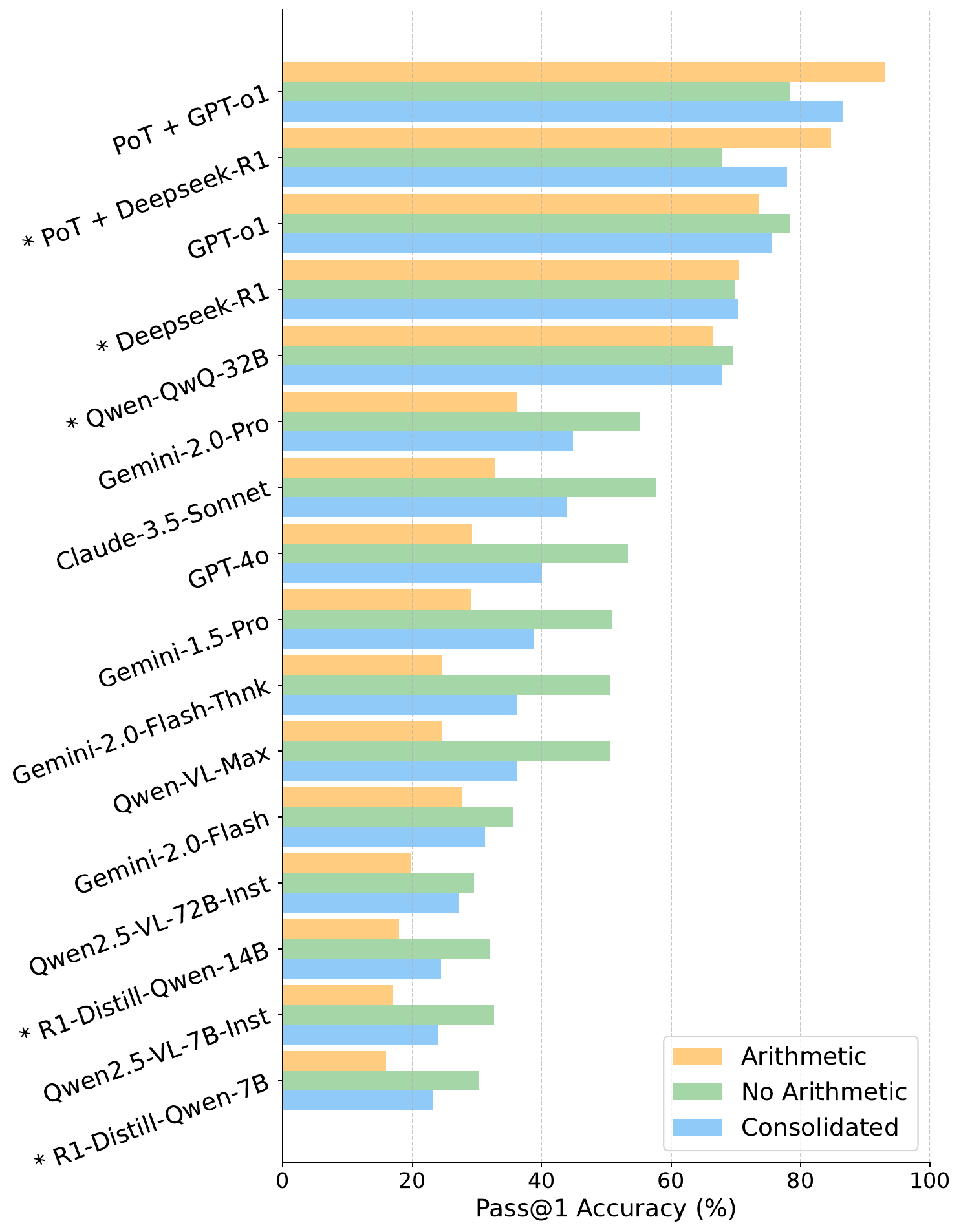}
        \caption{Performance of evaluated models on FAMMA-Basic. Each triplet of bars shows Pass@1 accuracy for the arithmetic subset (orange), the non-arithmetic subset (green), and their weighted consolidated score (blue). }
        \label{fig:leaderboard_basic}
    \end{minipage}%
    \hfill 
    \begin{minipage}[t]{0.47\textwidth}
        \centering
        \includegraphics[width=\textwidth]{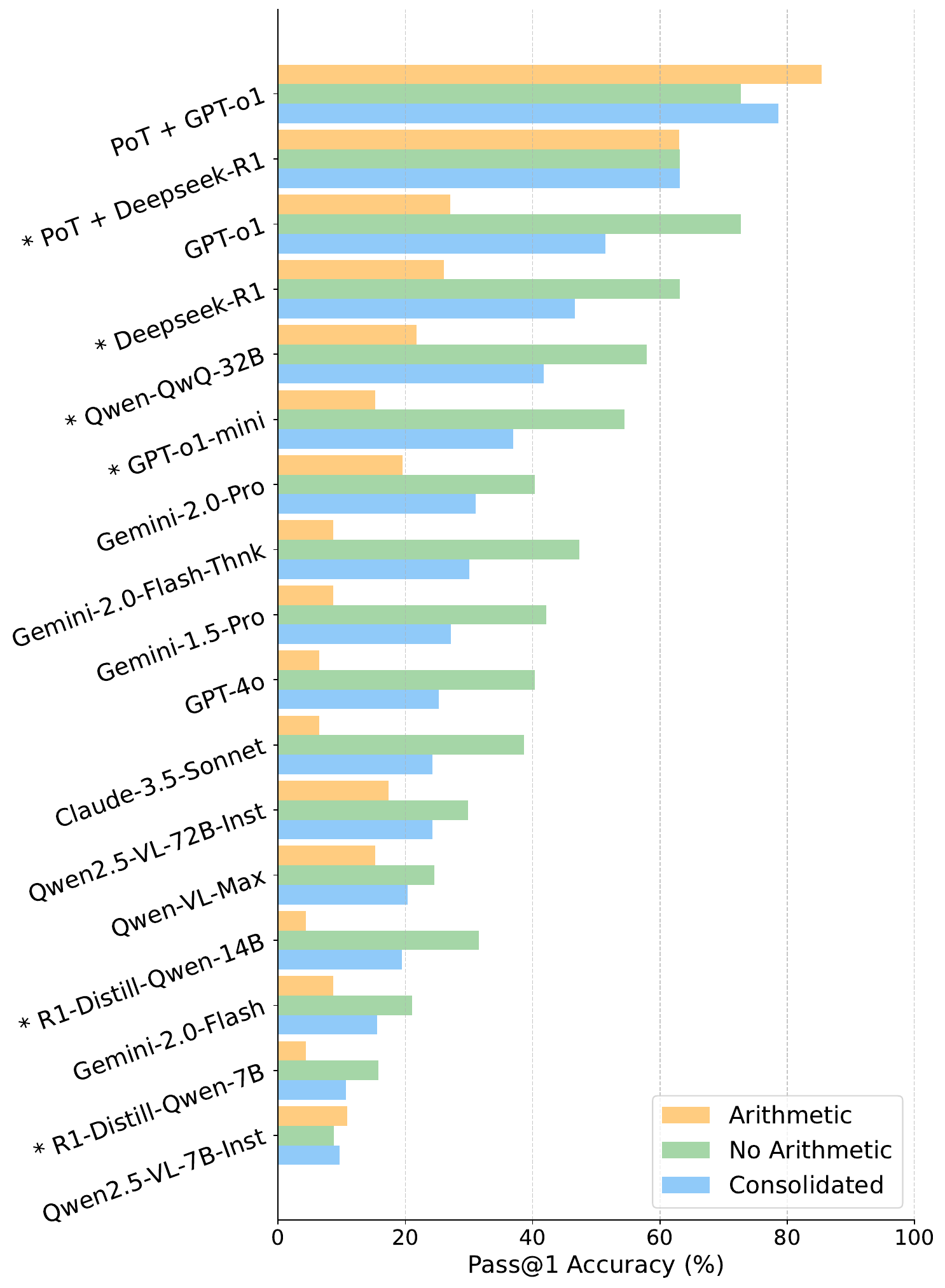}
        \caption{Performance of evaluated models on FAMMA-LivePro. Each bar group shows the same three metrics as in \cref{fig:leaderboard_basic}.}        
        \label{fig:leaderboard_livepro}
    \end{minipage}
    
\end{figure}

\begin{table*}[ht]
	\begin{center}
		\begin{small}
			\begin{sc}
				\begin{tabularx}{\textwidth}{l *{3}{C}*{3}{C} * {1}{C}}
					\toprule
					 & \multicolumn{3}{c}{Arithmetic Questions} & \multicolumn{3}{c}{Non-arithmetic Questions} & Overall \\
					\cmidrule(lr){2-4} \cmidrule(lr){5-7}
				Model	  & Easy & Medium & Hard &  Easy & Medium & Hard  &    \\
					\midrule
					PoT + GPT-o1 & $83.3$ & $92.8$ & $62.5$ & $95.4$ & $80.0$ & $38.8$& $78.6$ \\
                    GPT-o1 & $25.0$ & $35.7$ & $0.0$ & $95.4$ & $80.0$ & $38.8$& $51.4$ \\
                    * PoT + Deepseek-r1 & $70.1$ & $65.8$ & $20.4$ & $79.1$ & $66.6$ & $38.8$& $63.0$  \\
                    * Deepseek-r1 & $20.0$ & $35.7$ & $0.0$ & $79.1$ & $66.6$ & $38.8$& $46.6$ \\
					\bottomrule
				\end{tabularx}
			\end{sc}
		\end{small}
	\end{center}
	\caption{Pass@1 accuracy ($\%$) breakdown for the top-performing models on FAMMA–LivePro.}
	\label{tab:famma_live_details}
\end{table*}

\subsection{Results and Analysis}

\paragraph{Main results.} Overall Pass@1 scores of FAMMA-Basic and FAMMA-LivePro display in \cref{fig:leaderboard_basic} and \cref{fig:leaderboard_livepro}, respectively. We summarize the key findings as follows.

\begin{itemize}[leftmargin=*]
    \item FAMMA remains hard even for frontier models. On the contamination-free LivePro version, GPT-o1’s consolidated Pass@1 falls from $88\%$ on Basic to $60\%$, and DeepSeek-R1 drops from $80\%$ to $55\%$. Every model suffers a 25–40 percentage drop compared with Basic. In the hard tier specifically, seen in \cref{tab:famma_live_details}, GPT-o1 answers zero arithmetic questions and $38.8\%$ of the non-arithmetic questions, underscoring that even frontier models struggle.

    \item  Python tool-use markedly improves arithmetic performance. Shown in \cref{tab:famma_live_details},
with PoT prompting, GPT-o1’s arithmetic Pass@1 accuracy on LivePro jumps from $25.0 / 35.7 / 0 \%$ (easy / medium / hard) to $83.3 / 92.8 / 62.5 \%$, lifting its overall score to $78.6 \%$---the best recorded.  PoT likewise raises DeepSeek-R1’s arithmetic in similar scale. Structured tool-use therefore remains the most effective lever for numerical finance problems.

    \item Knowledge-heavy, non-arithmetic questions remain the bottleneck.
Even the best model GPT-o1 achieves only $38.8 \%$ on hard non-arithmetic items; most open-weight models hover near $25 \%$. Closing this gap will require advances in domain reasoning, not just numerical tool use.

\end{itemize}

\begin{figure}[t]
    \centering
    \begin{minipage}[t]{0.47\textwidth}
        \centering
        \includegraphics[width=\textwidth]{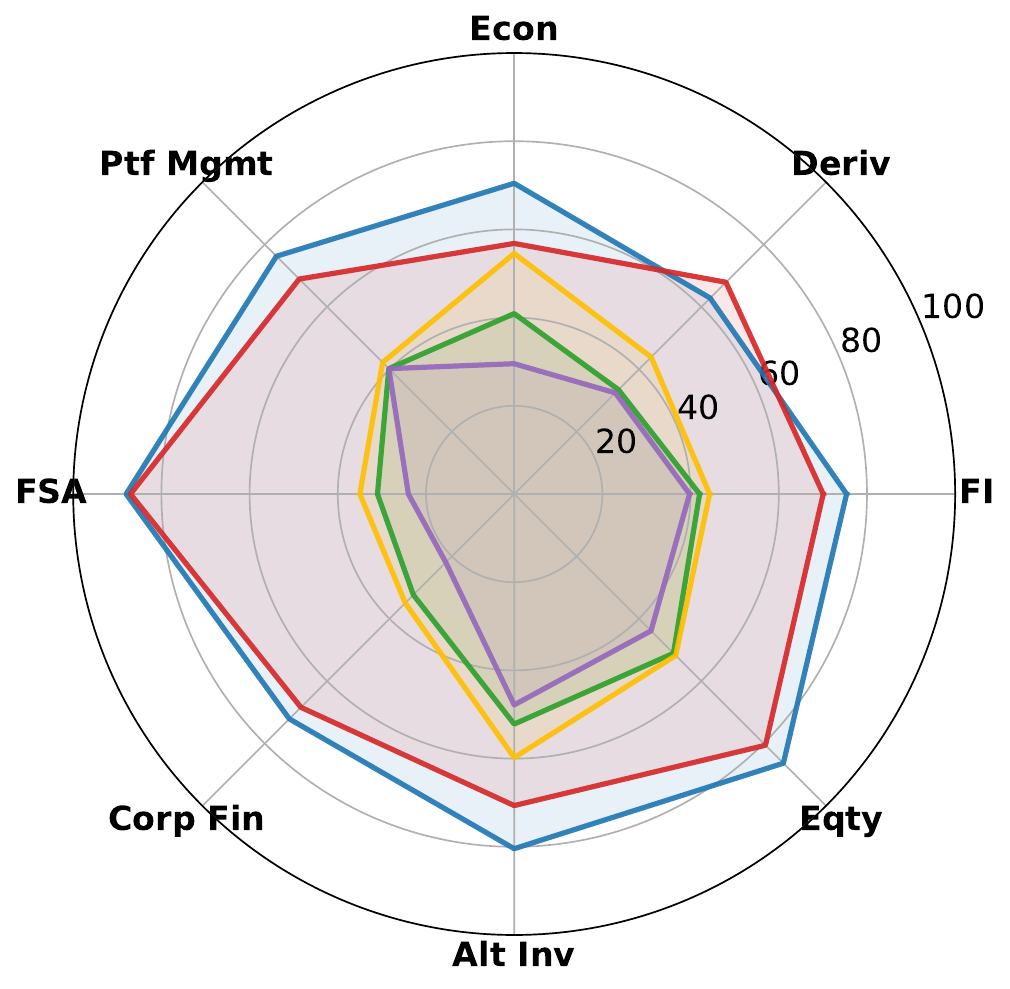}
        \caption{Pass@1 accuracy ($\%$) breakdown by subfields on FAMMA-Basic; the legend is identical to that in \cref{fig:subfield_pro}.}
        \label{fig:subfield_basic}
    \end{minipage}%
    \hfill 
    \begin{minipage}[t]{0.47\textwidth}
        \centering
        \includegraphics[width=\textwidth]{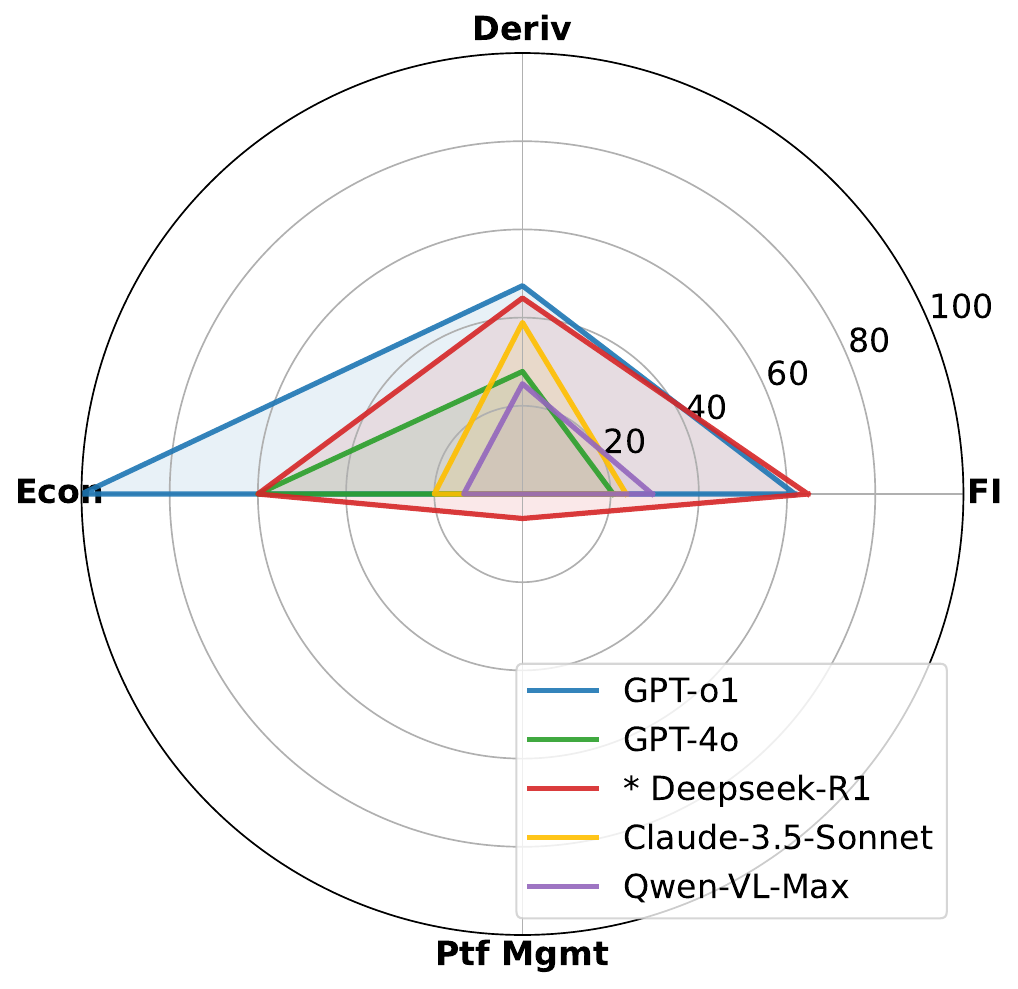}
        \caption{Pass@1 accuracy ($\%$) breakdown by subfields on FAMMA-LivePro.}        
        \label{fig:subfield_pro}
    \end{minipage}
    \vspace{-5mm}
\end{figure}

\paragraph{Analysis I: model performance across subfields.}
\cref{fig:subfield_basic} and \cref{fig:subfield_pro} give the Pass@1 breakdown by subfield for the three leading proprietary models and two strongest open-source models.

\begin{itemize}[leftmargin=*]
\item The widest gap between GPT-o1 and its strongest open-weight counterpart, DeepSeek-R1, appears in the Economics subfield on both the Basic and LivePro versions, indicating that DeepSeek-R1 remains comparatively weak in reasoning in economics context---an area where GPT models possibly benefit from better-curated pre-training data~\citep{quan2024econlogicqa}.
    \item On the Basic version, both GPT-o1 and Deepseek-R1 demonstrate the largest margin in financial statement analysis (FSA), whose context usually contains long tables (see \cref{fig:example_1}), against competitors, indicating their superior ability in table understanding and accounting knowledge.  
    \item In Portfolio Management subfiled, both GPT-o1 and DeepSeek-R1 perform respectably on Basic
($\approx 70\%$), yet their accuracy drops to $<5\%$ on
LivePro, underscoring that optimisation-heavy tasks remain virtually unsolved. 
\end{itemize}


\begin{figure}[t]
    \centering
    \begin{minipage}[t]{0.45\textwidth}
        \centering
        \includegraphics[width=\textwidth]{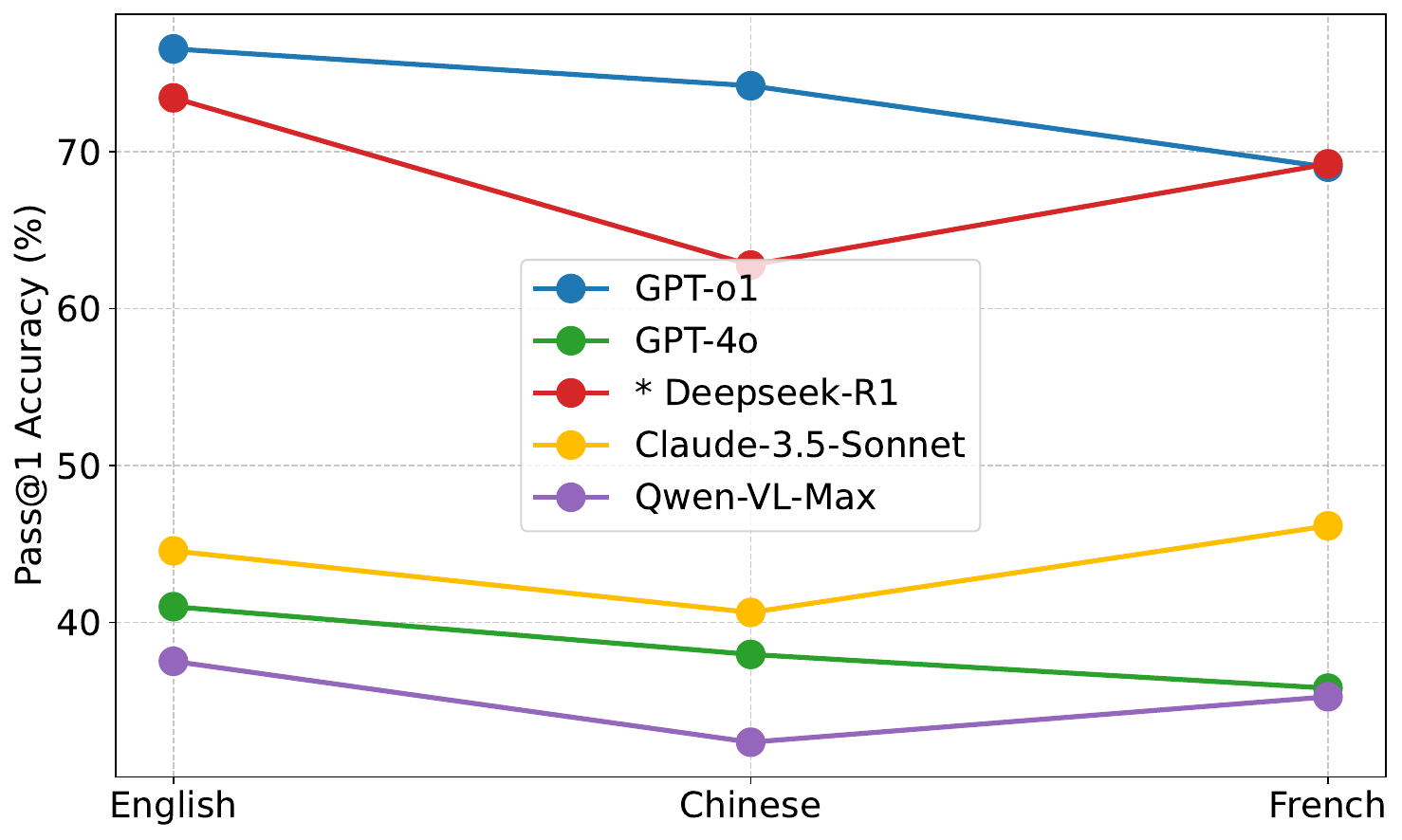}
        \caption{Pass@1 accuracy ($\%$) breakdown by language on FAMMA-Basic.}
        \label{fig:language_basic}
    \end{minipage}%
    \hfill 
    \begin{minipage}[t]{0.45\textwidth}
        \centering
        \includegraphics[width=\textwidth]{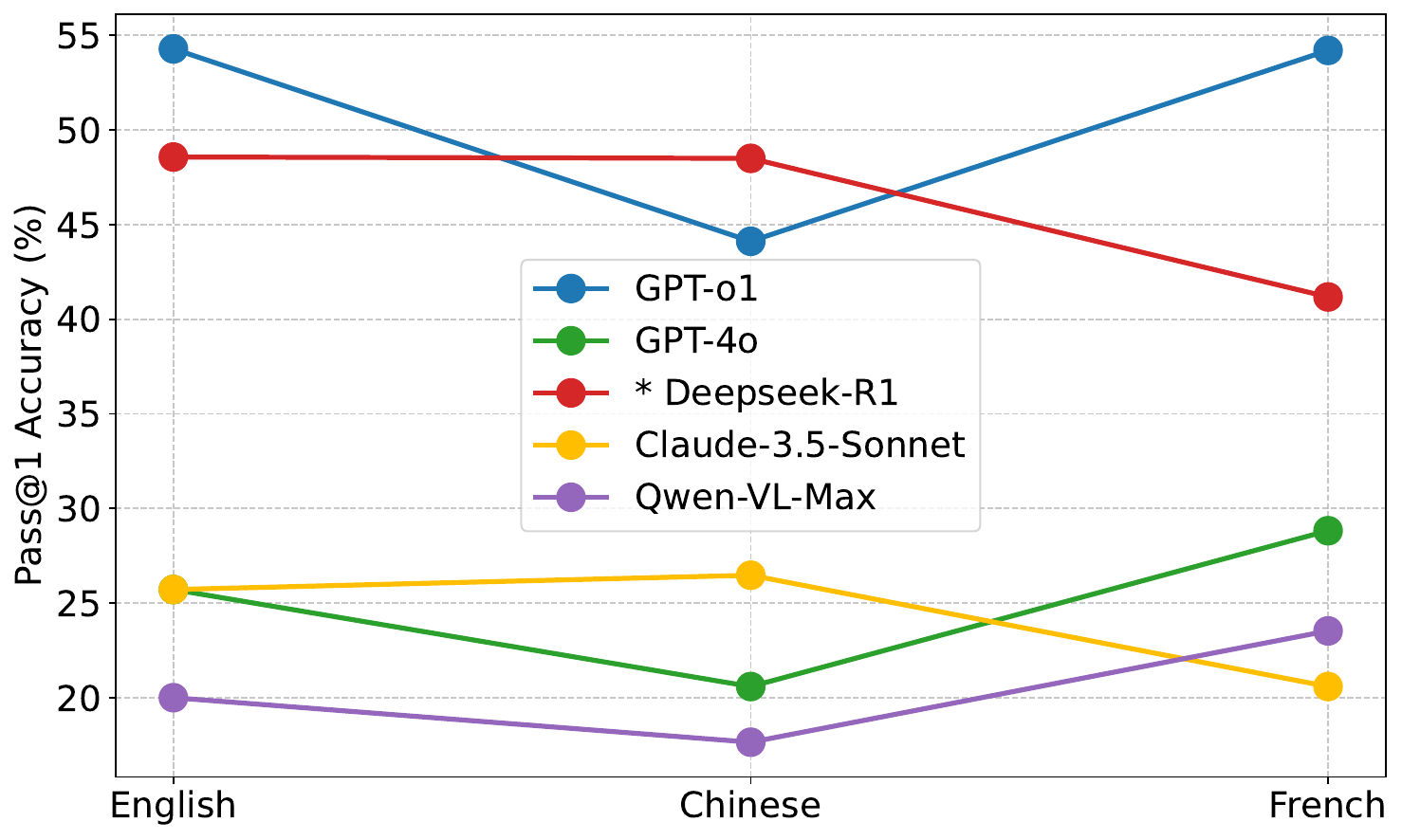}
        \caption{Pass@1 accuracy ($\%$) breakdown by language on FAMMA-LivePro.}        
        \label{fig:language_pro}
    \end{minipage}
    \vspace{-5mm}
\end{figure}

\paragraph{Analysis II: model performance across languages.} \cref{fig:language_basic} and \cref{fig:language_pro} reveal cross-lingual trends of models on Basic and LivePro. English is often strongest but not universally so: on LivePro, GPT-4o achieve its highest Pass@1 in French ($29\%$), and DeepSeek-R1 performs almost identically in English and Chinese ($48 \%$).  
Chinese remains the most challenging language: on Basic, four of the five models register their lowest accuracy in Chinese, and on LivePro this is still the case for three of the five. Conversely, French proves robust: on LivePro, GPT-o1, GPT-4o, and Claude-3.5 Sonnet match or even exceed their English accuracy, consistent with documentation indicating that these proprietary models undergo optimisation for French \citep{claudetechreport2023,claudetechreport2024}.

Overall, multilingual capability varies by model while Chinese remain the primary challenge.



\begin{figure}[t]
\begin{minipage}[t]{0.48\linewidth}
\begin{lstlisting}[caption={A response of GPT-o1 illustrating context misiterpretation: it mistakenly interprets the option-payoff diagram as a symmetric straddle, even though the actual payoff is asymmetric.},label={lst:err_cm}]
Consider the following payoff of an option strategy, <a payoff diagram consists of shorting three calls and one put>, how to replicate the payoff using calls and puts?
@
\vdots
@
The payoff shown peaks when the underlying price is at the strike. It is a short straddle which shorts @\sethlcolor{aired}\hl{one call and one put at strike 45.}@
\end{lstlisting}
\end{minipage}
\hfill
\begin{minipage}[t]{0.48\linewidth}
\begin{lstlisting}[caption={A response of GPT-o1 illustrating domain knowledge gap. A market maker facing strong upside pressure (deep bids / thin asks) usually wants to add inventory to profit from the likely price rise.},label={lst:main_err_dkg}]
Consider the following limit-order book of a specialist, <a table which shows considerable buying demaind and sparse limit sell orders>, should he increase to decrease his inventory? 
@
\vdots
@
@\sethlcolor{aired}\hl{Because buy orders are likely to push the price higher, the specialist would generally want to sell some of their holdings into this demand, thereby reducing inventory while prices are favorable.}@
\end{lstlisting}
\end{minipage}
\vspace{-5mm}
\end{figure}

\begin{figure}[t]
    \centering
    \begin{minipage}[t]{0.50\textwidth}
        \centering
        \includegraphics[width=\textwidth]{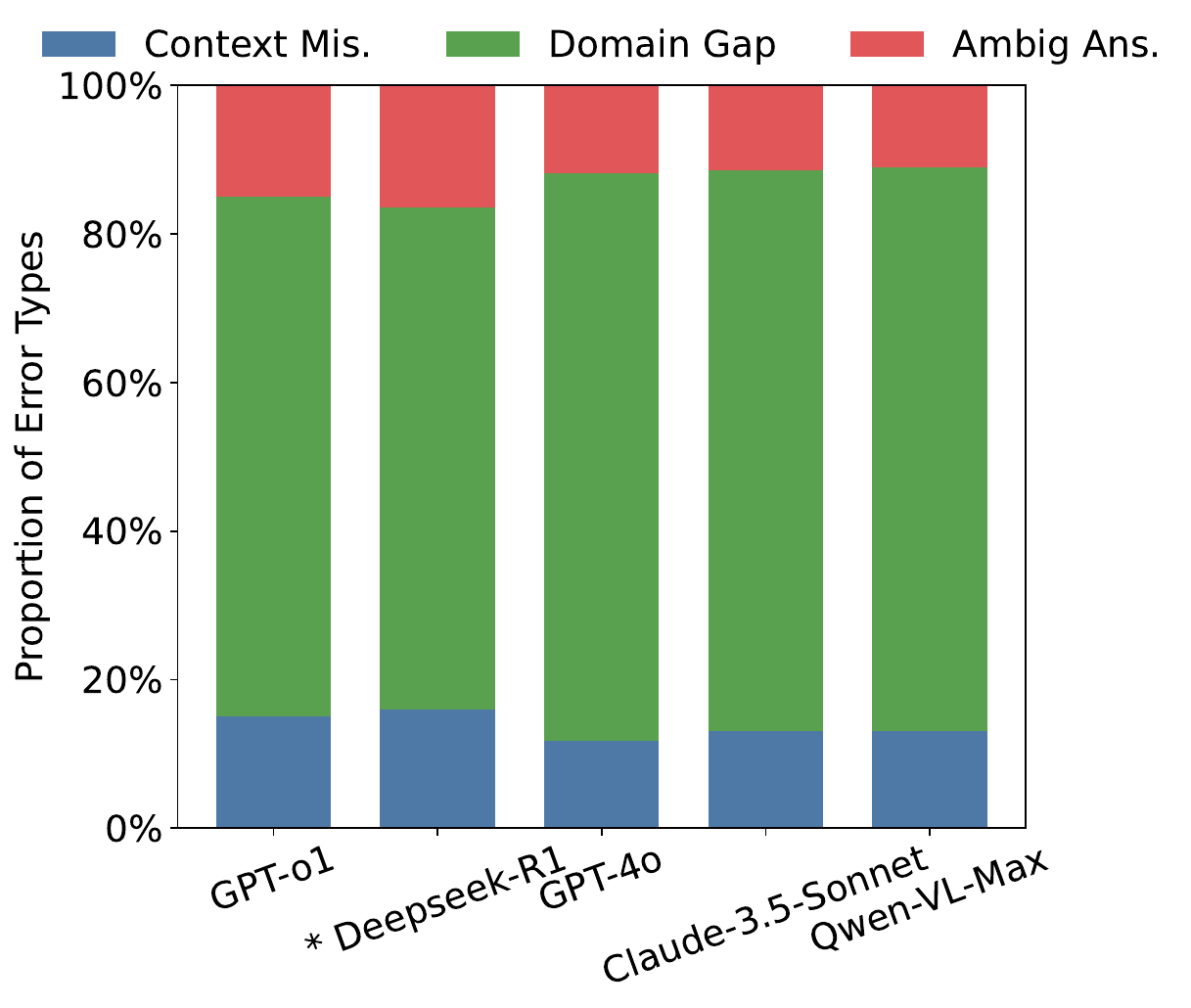}
        \caption{Error distribution on the non-arithmetic subset of FAMMA-LivePro, shown as each model’s error proportion normalized to its own total errors.}
        \label{fig:error_pro}
    \end{minipage}%
    \hfill 
    \begin{minipage}[t]{0.45\textwidth}
        \centering
        \includegraphics[width=\textwidth]{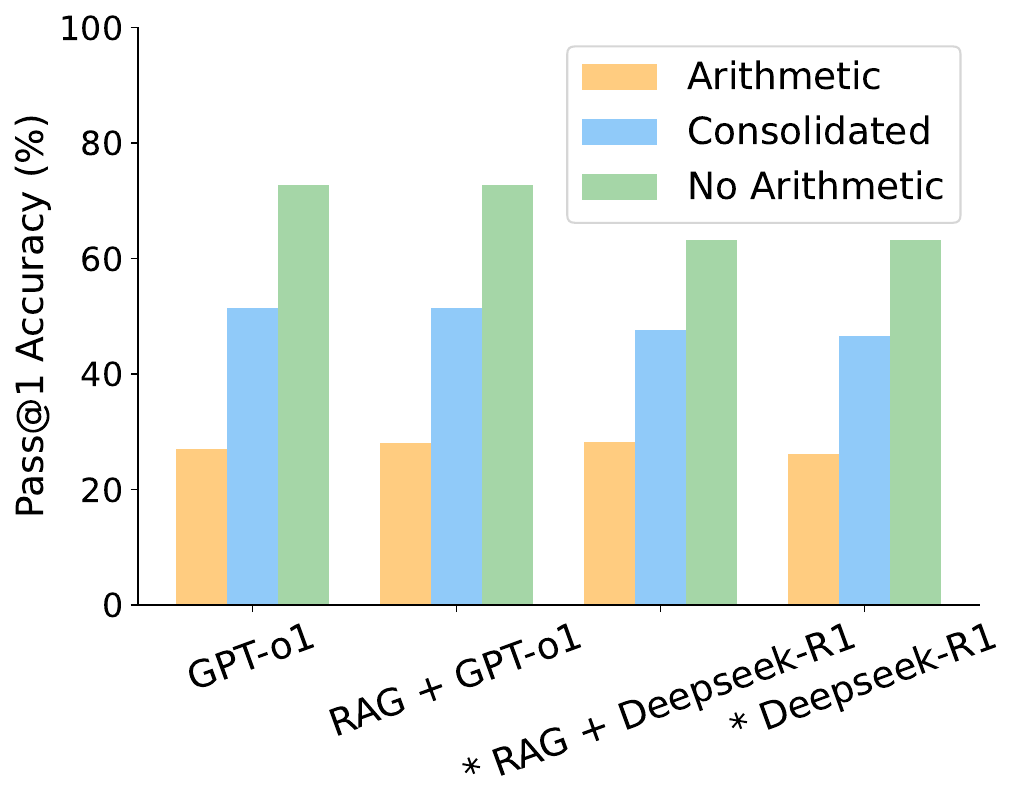}
        \caption{Pass@1 accuracy (\%) on FAMMA-LivePro for GPT-o1 and Deepseek-R1, with and without RAG.}        
        \label{fig:rag_pro}
    \end{minipage}
\end{figure}

\begin{figure}[t]
    \centering
    \begin{minipage}[t]{0.32\textwidth}
        \centering
        \includegraphics[width=\textwidth]{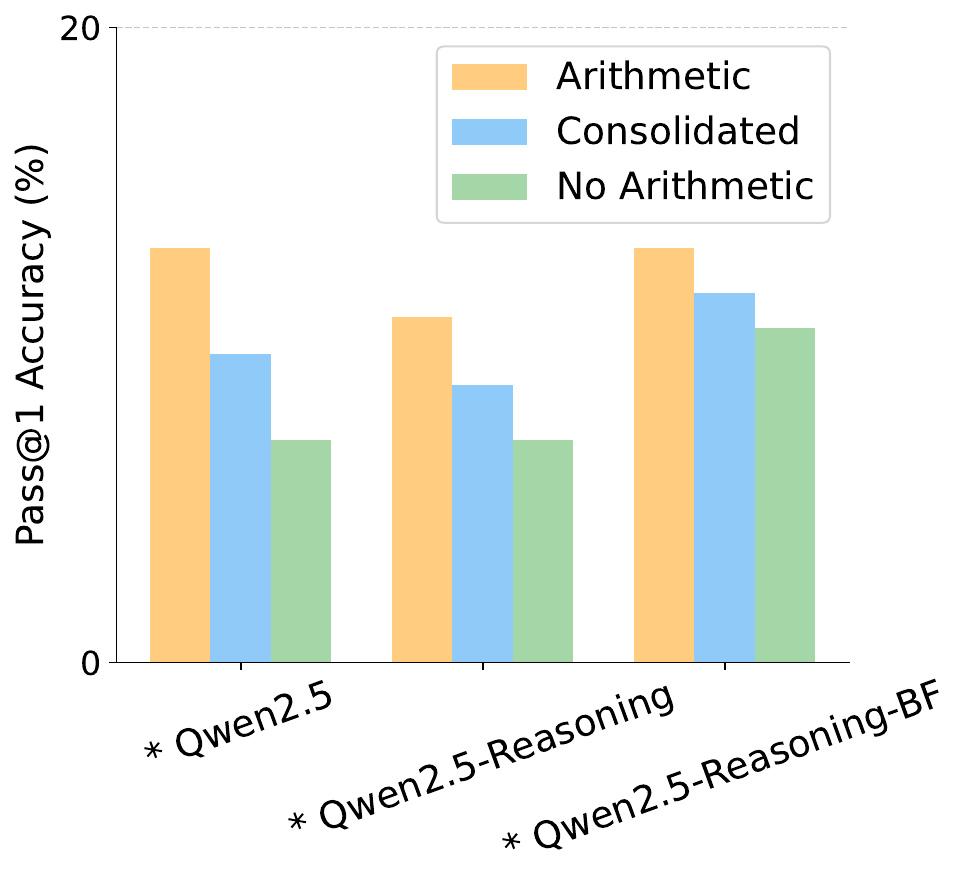}
        \caption{7B model performance on FAMMA-LivePro. BF = budget forcing.}
        \label{fig:distill_7b}
    \end{minipage}%
    \hfill 
    \begin{minipage}[t]{0.32\textwidth}
        \centering
        \includegraphics[width=\textwidth]{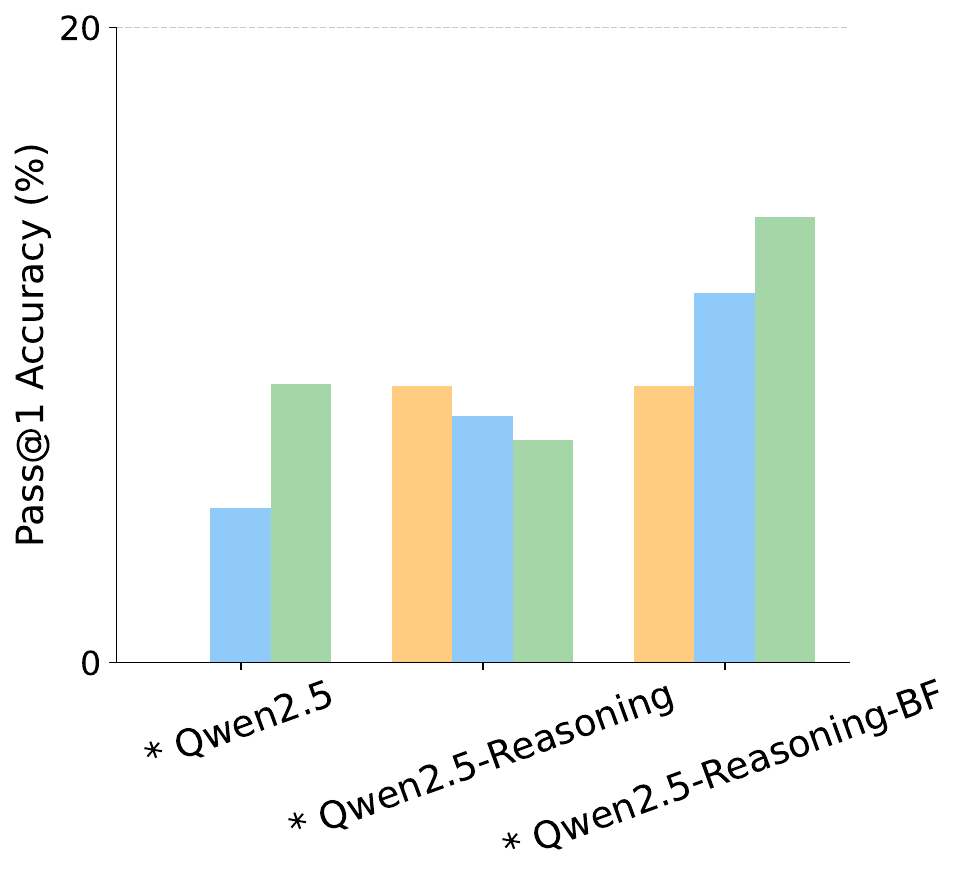}
        \caption{14B model performance on FAMMA-LivePro; the legend is identical to that in \cref{fig:distill_7b}.}        
        \label{fig:distill_14b}
    \end{minipage}
    \hfill 
    \begin{minipage}[t]{0.32\textwidth}
        \centering
        \includegraphics[width=\textwidth]{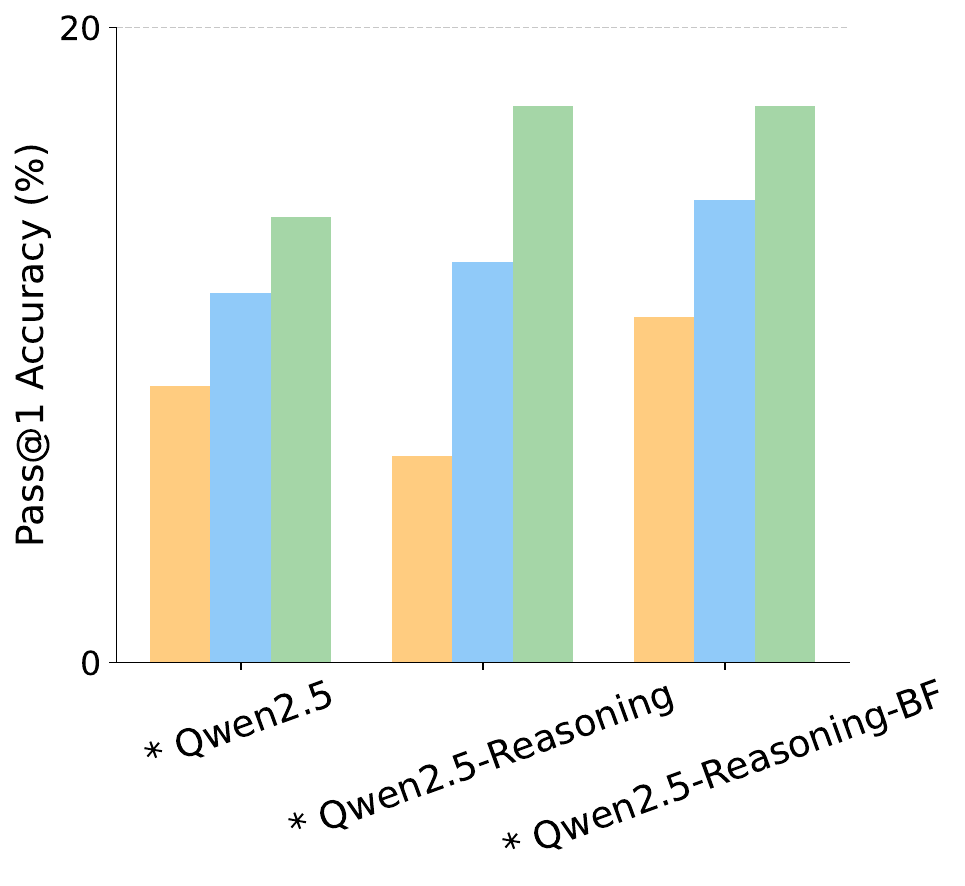}
        \caption{32B model performance on FAMMA-LivePro; the legend is identical to that in \cref{fig:distill_7b}.}        
        \label{fig:distill_32b}
    \end{minipage}
    \vspace{-5mm}
\end{figure}

\paragraph{Analysis III: error characterization.} Since PoT prompting largely resolves arithmetic questions, we restrict our subsequent error analysis to the non-arithmetic subset. 
We group errors into three types (see \cref{app:error} for the details of the pipeline): context misinterpretation, where the model misconstrues textual or visual inputs such as charts or tables (see \Cref{lst:err_cm}); domain-knowledge gaps, where it lacks the requisite financial expertise (\Cref{lst:main_err_dkg}); and ambiguous answers, where its reply is vague, incomplete, or misaligned with its own reasoning (\Cref{lst:main_err_aag} in \cref{app:error}). 

\cref{fig:error_pro} reveals that domain-knowledge gaps dominate the non-arithmetic subset of FAMMA-LivePro, making up roughly $75\%$ of every model’s errors. Context misinterpretations and ambiguous answers each contribute only a small fraction, indicating that further progress will reply mainly on deeper financial expertise.

\paragraph{Analysis V: can RAG help?} We apply retrieval-augmented generation (RAG) technique to GPT-o1 and Deepseek-r1 on LivePro.
See \cref{app:rag} for the implementation details of RAG.  
RAG left Pass@1 essentially unchanged. The likely reason is that these stronger base models already encode most of the classical textbook knowledge surfaced by retrieval, so the extra context offers little new signal and occasionally adds distraction.

\paragraph{Analysis VI: test-time scaling.} 
We perform supervised finetuning on Qwen2.5 series using FAMMA-reasoning data (see \cref{sec:reasoningdata}) to obtain our Qwen2.5-reasoning models. Furthermore, we employ budget forcing \citep{muennighoff2025s1} (BF), a simple decoding-time intervention by forcing a maximum and/or minimum number
of thinking tokens. Specifically, we ignore end-of-thinking token two times and append the string ``Wait'' (equivalent term in Chinese and French) to force it to continue reasoning when it tries to stop. Finetuning took one hour
on 8 NVIDIA H200 GPUs with PyTorch FSDP. The hyperparameters are outlined in \cref{app:training}.

Based on \cref{fig:distill_7b} -  \cref{fig:distill_32b}, we find:

\begin{itemize}[leftmargin=*]
    \item Finetuning with reasoning data (Qwen2.5-Reasoning) yields systematic improvements over the base model on the FAMMA-LivePro. The gains are modest for the 7B and 14B, but grow with scale: the 32B model records an increase ($\approx 4\%$) in Pass@1 on the non-arithmetic subset.  
    \item BF provides a complementary gain. 
    Applied to Qwen2.5-Reasoning variants, BF increases overall Pass@1 by $\approx 3\%/4\%/4\%$ for the 7B, 14B, and 32B models, respectively. The effect scales favourably with model size: 14B BF model brings $5\%$ on non-arithmetic questions while 32B BF model gains additional $3\%$ on arithmetic questions. 
\end{itemize}



To conclude, finetuing with reasoning trajectories largely improve the performance on LivePro. Equipping the reasoning model with the simple budget forcing technique operates in a complementary scaling paradigm.

\section{Conclusion}
We present a multilingual, multimodal benchmark to advance robust financial QA systems. Our benchmark not only lays the groundwork for evaluating existing models but also aims to accelerate advancements in this field by providing a set of high-quality dataset.




\clearpage
\bibliographystyle{icml2020_url}
\bibliography{references} 

\clearpage
\appendix
\appendixpage

\section{Dataset Details}
\label{app:dataset}

\paragraph{FAMMA-Basic collection.}  The question-response pairs are primarily collected from free online resources, quizzes, textbooks, and other study materials. See \cref{tab:data_source} for more details.

\paragraph{FAMMA-LivePro collection.}

The FAMMA-LivePro subset was developed in close collaboration with four senior practitioners in quantitative finance.  
Although they have elected to remain anonymous, their backgrounds and specific contributions are summarised below.

\begin{itemize}[leftmargin=*]
    \item Expert~A (Diplôme d'Ing\'enieur, \'Ecole Polytechnique; DEA d'El Karoui) spent over a decade as an options trader at leading hedge funds and contributed \emph{Question 7}.  
    \item Expert~B (M.~Sc.\ Applied Mathematics, \'Ecole Polytechnique; DEA d'El Karoui) brings 10 years of quantitative-research and generative-AI experience in large technology firms and authored \emph{Questions 3-6}.  
    \item Expert~C (M.~Sc.\ Finance, Shanghai University of Finance \& Economics) has more than a decade of portfolio-management practice and prepared \emph{Questions 1-2}.  
    \item Expert~D (Ph.D.\ Mathematics, University of Science and Technology of China, specialising in stochastic control and optimisation) performed a full technical validation of all items.
\end{itemize}

\begin{table*}[tb]
	\begin{center}
		\begin{small}
			\begin{sc}
                    \renewcommand{\arraystretch}{1.3}
                    \setlength{\tabcolsep}{12pt}
				\begin{tabularx}{\textwidth} 
                    {l X X }
					\toprule
                    Materials  & Language & Source \\
                    \midrule
                    Quizzes on finance-related courses   &   English &
 \href{https://ocw.mit.edu/courses}{MIT OpenCourse} \\
                    Textbook: Numerical Probability   &   English & \href{https://math.uni.lu/thalmaier/Stud_Proj/Pages_Gilles_Numerical_Probability.pdf}{Online PDF} \\
                    Textbook: Paul Wilmott on Quantitative Finance   &   English & \href{https://github.com/ReDxDaGer/quant-resources/tree/main/Books}{Github} \\
                    Quizzes on finance-related courses    &   Chinese & \href{https://bbs.pinggu.org/}{Renming University} \\
                    Quizzes on finance-related courses    &   French & \href{https://fr.scribd.com/}{Scribd} \\
                    Quizzes on finance-related courses    &   French & \href{https://www.academia.edu/}{Academia} \\
					\bottomrule
				\end{tabularx}
			\end{sc}
		\end{small}
	\end{center}
	\caption{Selected sources as references for generating questions. }
	\label{tab:data_source}
\end{table*}

\paragraph{Data format.} Following data validation, we provide the following information for each question:
\begin{itemize}[leftmargin=*]

\item Question ID: a unique identifier for the question across the whole dataset.
\item Context: relevant background information related to the question.
\item Question: the specific query being asked.
\item Images: directories of images referenced in the context or question.
\item Options: a list of possible answers, applicable only to multiple-choice questions.
\item Question type: categorized as either multiple-choice or open-ended.
\item Main question ID: a unique identifier for the question within its context; questions with the same context share the same ID.
\item Sub question ID: a unique identifier for the question within its corresponding main question.
\item Answer: a concise and accurate response.
\item Explanation: a detailed justification for the answer.
\item Images for explanation: directories of images supporting the explanation.
\item Subfield: the specific area of expertise to which the question belongs, categorized into eight subfields.
\item Language: the language in which the question text is written.
\item Difficulty: a measure of the question's complexity based on the level of reasoning required.
\item is$\_$arithmetic: whether the question is an arithmetic question that needs heavy calculation.
\end{itemize}




\begin{figure}[t]
\begin{minipage}[t]{0.5\linewidth}
\begin{lstlisting}[caption={Multi-choice questions in JSON representation.},label={lst:mc_question_json_format}]
{
    "question_id": "English_validation_86",
    "context": "The following data are available relating to the performance of Wildcat Fund and the market portfolio: <image_1>",
    "question": "The risk-free return during the sample period was 7%. Calculate Sharpe's measure of performance for Wildcat Fund.",
    "options": "['1.00%', '8.80%', '44.00%', '50.00%']",
    "image_1": "/9j/4AAQSkZJRgABAQAAAQABAAD...]",
    "image_2": null,
    "image_3": null,
    "image_4": null,
    "image_5": null,
    "image_6": null,
    "image_7": null,
    "image_type": "table",
    "answers": "C",
    "explanation": "(18 - 7)/25 = .44.",
    "topic_difficulty": "easy",
    "question_type": "multiple-choice",
    "subfield": "portfolio management",
    "language": "english",
    "main_question_id": 369,
    "sub_question_id": 2,
    "ans_image_1": null,
    "ans_image_2": null,
    "ans_image_3": null
}
\end{lstlisting}
\end{minipage}
\begin{minipage}[t]{0.5\linewidth}
\begin{lstlisting}[caption={Open questions in JSON representation.},label={lst:open_question_json_format}]
 {
        "question_id": "English_validation_42",
        "context": "Cleveland Compressor and Pnew York Pneumatic are competing manufacturing firms. Their financial statements are printed here.<image_1><image_2><image_3><image_4>",
        "question": "Which firm has the larger investment in current assets? Why?",
        "options": "",
        "image_1": "/9j/4AAQSkZJRgABAQAAAQABAAD...",
        "image_2": "/9j/4AAQSkZJRgABAQAAAQABAAD...",
        "image_3": "/9j/4AAQSkZJRgABAQAAAQABAAD...",
        "image_4": "/9j/4AAQSkZJRgABAQAAAQABAAD...",
        "image_5": null,
        "image_6": null,
        "image_7": null,
        "image_type": "table",
        "answers": "Cleveland Compressor.",
        "explanation": "Cleveland Compressor holds the larger investment in current assets. It has current assets of $92,616 while Pnew York Pneumatic has $70,101 in current assets. The main reason for the difference is the larger sales of Cleveland Compressor.",
        "topic_difficulty": "hard",
        "question_type": "open question",
        "subfield": "financial statement analysis",
        "language": "english",
        "main_question_id": 329,
        "sub_question_id": 3,
        "ans_image_1": null,
        "ans_image_2": null,
        "ans_image_3": null
    },
\end{lstlisting}
\end{minipage}
\end{figure}

\section{Copyright and Licensing}
\label{app:coppyright}

\paragraph{FAMMA-Basic copyright.}
While FAMMA-Basic is constructed from publicly available educational sources, we believe its use qualifies as fair use under U.S. copyright law, satisfying all four criteria defined by the U.S. Copyright Office.\footnote{\small \url{https://www.copyright.gov/fair-use/}} Below, we address each factor in turn:
\begin{itemize}[leftmargin=*]
\item Purpose and character of use. FAMMA-Basic is intended solely for non-commercial academic research. All questions have been substantially rewritten, expanded, and embedded into novel test formats tailored for evaluating AI systems—constituting a clear transformative use that differs fundamentally from the original educational purposes.
\item Nature of the copyrighted work.
The source material consists predominantly of factual and technical content (e.g., financial concepts, valuation methods, market structures) originally published for educational purposes. These works are factual in nature, designed for open knowledge dissemination, and in many cases publicly accessible.
\item Amount and substantiality. FAMMA-Basic includes 1,975 questions, drawn from a wide range of sources. No single source contributes more than $5\%$ of the total dataset. All items have been filtered and rewritten to avoid direct reuse of key or proprietary material, preserving only general ideas or thematic inspiration.
\item Effect on the market.
The dataset does not compete with, substitute for, or diminish the market value of the original educational resources. Its target audience, AI researchers, is different from the original audience of finance students or instructors. The dataset has no commercial use or monetization intent.
\item Source attribution.
The primary sources include MIT OpenCourseWare (which permits redistribution and modification) and materials hosted on Academia.edu (which fall under limited fair use). We have carefully cited all source materials, with a detailed breakdown provided in \cref{tab:data_source}.

\end{itemize}

\paragraph{FAMMA–LivePro copyright.}
All contributions to FAMMA–LivePro were obtained with written consent from the invited experts, who granted permission for their use in this research.

\section{Experiment Details}
\label{app:exp}

\subsection{Details on LLM-powered evaluation.}
\label{app:llm_eval}

The LM evaluator is instructed to understand the format and structure of the questions, as well as to consider the key points in the ground-truth answers for evaluating the responses. Please refer to \Cref{lst:judge_format} in \cref{app:exp} for the instructions provided for evaluating the answers. Note that for open-ended questions, where both gold and generated answers are provided, there is a single correct answer, making the 1-0 correctness straightforward to determine. We set the temperature of the LM evaluator as $0$ to keep the evaluation results deterministic.

\subsection{Details on error characterization.}
\label{app:error}

These instances are firstly categorized by LM-evaluators (see \Cref{lst:analysis_format} in \cref{app:exp} for the instruction prompt), then validated by human expert based on their knowledge and the golden explanations if available.

\begin{figure}[t]
\begin{minipage}[t]{0.5\linewidth}
\begin{lstlisting}[caption={A sample case response of GPT-4o due to ambiguous answer generation: it correctly performs the computation but reaches to a wrong final result.},label={lst:main_err_aag}]
@
\vdots
@
The expected holding-period return (HPR) is calculated by multiplying each possible return by its probability and summing the results. 
 The calculation is as follows: (0.30 * 18%) + (0.50 * 12%) + (0.20 * -5%) = 5.4% + 6% - 1% = 10.4%. Therefore, the expected HPR for KMP stock is @\sethlcolor{aired}\hl{10.88\%}@.
\end{lstlisting}
\end{minipage}
\begin{minipage}[t]{0.5\linewidth}
\end{minipage}
\end{figure}

\subsection{Details on RAG experiments}
\label{app:rag}

\paragraph{RAG setup.}
We utilize 5 CFA Level III curriculum textbooks---``CFA Level III SchweserNotes, Books 1-5, 2023'', which comprehensively cover most of the topics found in FAMMA--- as the external knowledge source. The notes are in PDF format, each consisting of 200-300 pages with quizzes at the end of every chapter, though these quizzes are not included in FAMMA. 

To manage costs, we implemented a customized RAG system using GTE-Multilingual-Base~\citep{zhang2024mgte}, a 305M-parameter model optimized for multilingual text processing. The top 5 results were retrieved based on cosine similarity. Generation prompt: the retrieved contexts were appended to the question using the template described in \Cref{lst:mc_instruction_format} and \Cref{lst:open_instruction_format}, followed by the instruction: "Answer the question based on the provided information." This approach balances cost-efficiency with strong multilingual capabilities.

\begin{figure}[t]
\begin{minipage}[t]{0.5\linewidth}
\begin{lstlisting}[caption={Format of our instruction prompt on multi-choice questions.},label={lst:mc_instruction_format}]
You are a highly knowledgeable financial expert. Please answer multiple-choice questions in the finance domain. You are given context, images, questions and options.
The questions are multilingual (either in English, Chinese, or French) and multimodal (containing images as part of the question). '<image_1>, <image_2> ...' mentioned in the text of the context or question are sequential placeholders for images, which are fed at the same time as the textual information. 
If no image information is provided, you must answer based solely on the given information.
Besides, the question may contain several sub-questions that share the same context, and the answer to each sub-question may depend on the answers to previous ones.
The question format is

context: <context>
sub-question-1: <sub-question-1>
sub-question-2: <sub-question-2>
sub-question-3: <sub-question-3>
...

Now consider the following question:
context: {context} 
{sub_questions}

Please provide the chosen answer and a precise, detailed explanation of why this choice is correct. The explanation should be in the same language as the question and should not exceed 400 words.
Your response must be in a standard JSON format:
{{
   sub-question-1: {{
       answer-1: <answer-1>,
       explanation-1: <explanation-1>
   }},
   sub-question-2: {{
       answer-2: <answer-2>,
       explanation-2: <explanation-2>
   }},
   sub-question-3: {{
       answer-3: <answer-3>,
       explanation-3: <explanation-3>
   }},
   ...
}}
Ensure that the response strictly adheres to JSON syntax without any additional content.  
\end{lstlisting}
\end{minipage}
\begin{minipage}[t]{0.5\linewidth}
\begin{lstlisting}[caption={Format of our instruction prompt on open questions.},label={lst:open_instruction_format}]
You are a highly knowledgeable financial expert. Please answer open-ended questions in the finance domain. 
The questions are multilingual (either in English, Chinese, or French) and multimodal (containing images as part of the question). '<image_1>, <image_2> ...' mentioned in the text of the context or question are sequential placeholders for images, which are fed at the same time as the textual information. 
If no image information is provided, you must answer based solely on the given information.
Besides, the question may contain several sub-questions that share the same context, and the answer to each sub-question may depend on the answers to previous ones.
The question format is

context: <context>
sub-question-1: <sub-question-1>
sub-question-2: <sub-question-2>
sub-question-3: <sub-question-3>
...

Now consider the following question:
context: {context} 
{sub_questions} 

Please provide the answer and a precise, detailed explanation. The explanation should be in the same language as the question and should not exceed 400 words.
Your answer must be in a standard JSON format:
{{
   sub-question-1: {{
       answer-1: "answer-1",
       explanation-1: "explanation-1"
   }},
   sub-question-2: {{
       answer-2: "answer-2",
       explanation-2: "explanation-2"
   }},
   sub-question-3: {{
       answer-3: "answer-3",
       explanation-3: "explanation-3"
   }},
   ...
}}
Ensure that the response strictly adheres to JSON syntax without any additional content. 
\end{lstlisting}
\end{minipage}
\end{figure}

\begin{figure}[t]
\begin{minipage}[t]{0.5\linewidth}
\begin{lstlisting}[caption={Format of our prompt on judging the correctness of the model output.},label={lst:judge_format}]
You are a highly knowledgeable expert and teacher in the finance domain. 
You are reviewing a student's answers to financial questions. 
The questions are multilingual (either in English, Chinese, or French) and multimodal (containing images as part of the question). '<image_1>, <image_2> ...' mentioned in the text of the context or question are sequential placeholders for images, which are fed at the same time as the textual information.
You are given the context, the question, the student's answer and the student's explanation and the ground-truth answer. 
Please use the given information and refer to the ground-truth answer to determine if the student's answer is correct.

The input information is as follows:

context: {context}
question: {question}
student's answer: {model_answer}
student's explanation: {model_explanation}
ground-truth answer: {answer}
 
Please respond directly as either 'correct' or 'incorrect'. 
\end{lstlisting}
\end{minipage}
\begin{minipage}[t]{0.5\linewidth}
\begin{lstlisting}[caption={Format of our prompt on error analysis on model's output.},label={lst:analysis_format}]
You are a highly skilled expert in error analysis for AI models in the finance domain. You are reviewing collected incorrect answers to financial questions.
The questions are multilingual (either in English, Chinese, or French) and multimodal (containing images as part of the question). '<image_1>, <image_2> ...' mentioned in the text of the context or question are sequential placeholders for images, which are fed at the same time as the textual information.
You are given the context, the question, the student's answer, the student's explanation and the ground-truth. 

You need to classify these incorrect answers based on the provided categories: perceptual errors, lack of knowledge, reasoning errors, and other errors. Here are the definitions for each error type:

Data misinterpretation: .(omitted)
Incomplete context understanding: ...(omitted)
Domain knowledge gaps:  ...(omitted)
Ambiguous answer generation:  ...(omitted)

The input is as follows; use these details to determine the primary error category.

context: {context}
question: {question}
model incorrect answer: {model_answer}
model explanation: {model_explanation}
ground-truth answer: {answer}

Now please provide the result directly, identifying the error category as one of: data misinterpretation, incomplete context understanding, domain knowledge gaps, or ambiguous answer generation.
\end{lstlisting}
\end{minipage}
\end{figure}

\subsection{Details on training Qwen2.5 series.}
\label{app:training}
We take a series of Qwen2.5 that has already been pretrained and instruction tuned and further finetune it for reasoning. We use token delimiters to separate the thinking stage from the answering stage. We enclose the thinking stage with $<$think$>$ and $<$answer$>$.

For 7B/14B/32B variants, we perform the LoRA fine-tune with the identical hyperparameters. Each model is trained for five epochs with an effective batch size of 64 (per-device batch size = 1, gradient accumulation = 8) on eight NVIDIA H200 GPUs. We use adapters with rank=8 and $\alpha=32$, bfloat16 precision, and a base learning rate of $e^{-5}$ scheduled by cosine decay. The loss is computed only on reasoning traces and final solutions, excluding the question tokens. A full run completes in approximately one hour.

\section{Future Work}
Looking ahead, future work can build on our findings and the established benchmark in several important ways:
\begin{itemize}[leftmargin=*]
    \item Dataset enrichment: we propose further enhancement of the benchmark by incorporating a wider array of languages, and richer multimodal content. This approach will aim to capture the intricacies of real-world financial inquiries more effectively.

    \item we encourage researchers to develop innovative models using refined RAG techniques~\citep{langchain,xue2023dbgpt} or advanced prompting methods~\citep{gpt4o1}, to compete with those featured on the leaderboard, promoting healthy competition and further advancements.

    \item Real-world applications: integrating our benchmark into practical applications can provide valuable insights into its effectiveness in real-world contexts, such as improving customer service chatbots in financial institutions or refining automated advisory systems.
\end{itemize}

\section{Limitations}\label{app:limit}
The automatic evaluation relies on state-of-the-art LLMs (e.g., GPT-4o) and therefore inherits their residual imperfections—chiefly occasional hallucinations. As more capable and cost-efficient models become available, we expect the reliability of this evaluation pipeline to improve correspondingly. 

\section{Broader impacts}\label{app:impact}
By furnishing an open, multilingual and multimodal benchmark for financial QA, FAMMA lowers the entry barrier for both academic and industrial groups seeking to measure domain-specific reasoning in large language models.

\section{Safeguards} \label{app:safe}
The dataset is distributed under a research-only license that prohibits commercial use. All source documents were screened to remove personally identifiable information and sensitive customer data.

\section{Approvals for research with human subjects} \label{app:human}
Participants (expert annotators) were given an information sheet detailing the study’s purpose, their voluntary involvement, the absence of personal-data collection, and the planned public release of the benchmark. They work without compensation. Written consent was obtained. They could withdraw at any time, and no identifying information is retained. Because our institution lacks a formal IRB pathway for low-risk studies, we followed an internal peer-led ethics review, consistent with the NeurIPS guideline that an equivalent informal process may be used when no formal mechanism exists.


\clearpage
\newpage
\section*{NeurIPS Paper Checklist}

The checklist is designed to encourage best practices for responsible machine learning research, addressing issues of reproducibility, transparency, research ethics, and societal impact. Do not remove the checklist: {\bf The papers not including the checklist will be desk rejected.} The checklist should follow the references and follow the (optional) supplemental material.  The checklist does NOT count towards the page
limit. 

Please read the checklist guidelines carefully for information on how to answer these questions. For each question in the checklist:
\begin{itemize}
    \item You should answer \answerYes{}, \answerNo{}, or \answerNA{}.
    \item \answerNA{} means either that the question is Not Applicable for that particular paper or the relevant information is Not Available.
    \item Please provide a short (1–2 sentence) justification right after your answer (even for NA). 
\end{itemize}

{\bf The checklist answers are an integral part of your paper submission.} They are visible to the reviewers, area chairs, senior area chairs, and ethics reviewers. You will be asked to also include it (after eventual revisions) with the final version of your paper, and its final version will be published with the paper.

The reviewers of your paper will be asked to use the checklist as one of the factors in their evaluation. While "\answerYes{}" is generally preferable to "\answerNo{}", it is perfectly acceptable to answer "\answerNo{}" provided a proper justification is given (e.g., "error bars are not reported because it would be too computationally expensive" or "we were unable to find the license for the dataset we used"). In general, answering "\answerNo{}" or "\answerNA{}" is not grounds for rejection. While the questions are phrased in a binary way, we acknowledge that the true answer is often more nuanced, so please just use your best judgment and write a justification to elaborate. All supporting evidence can appear either in the main paper or the supplemental material, provided in appendix. If you answer \answerYes{} to a question, in the justification please point to the section(s) where related material for the question can be found.

IMPORTANT, please:
\begin{itemize}
    \item {\bf Delete this instruction block, but keep the section heading ``NeurIPS Paper Checklist"},
    \item  {\bf Keep the checklist subsection headings, questions/answers and guidelines below.}
    \item {\bf Do not modify the questions and only use the provided macros for your answers}.
\end{itemize}


\begin{enumerate}

\item {\bf Claims}
    \item[] Question: Do the main claims made in the abstract and introduction accurately reflect the paper's contributions and scope?
    \item[] Answer: \answerYes{} 
    \item[] Justification: the abstract and the last paragraph of \cref{sec:intro} are consistently and accurately articulate the paper’s principal claims.

\item {\bf Limitations}
    \item[] Question: Does the paper discuss the limitations of the work performed by the authors?
    \item[] Answer: \answerYes{} 
    \item[] Justification: see \cref{app:limit}.

\item {\bf Theory assumptions and proofs}
    \item[] Question: For each theoretical result, does the paper provide the full set of assumptions and a complete (and correct) proof?
    \item[] Answer: \answerNA{}. 
    \item[] Justification: our paper has no theoretical part that needs a set of assumptions and proof.

    \item {\bf Experimental result reproducibility}
    \item[] Question: Does the paper fully disclose all the information needed to reproduce the main experimental results of the paper to the extent that it affects the main claims and/or conclusions of the paper (regardless of whether the code and data are provided or not)?
    \item[] Answer: \answerYes{} 
    \item[] Justification: The evaluation details are discussed in ``Lm-powered evaluation'' paragraph in \cref{sec:exp_setup} and \cref{app:llm_eval}. The 
    training details of reasoning models can be found in \cref{app:training}. The code has been released at \small 
 \url{https://github.com/famma-bench/bench-script}.

\item {\bf Open access to data and code}
    \item[] Question: Does the paper provide open access to the data and code, with sufficient instructions to faithfully reproduce the main experimental results, as described in supplemental material?
    \item[] Answer: \answerYes{} 
    \item[] Justification: The dataset is publicly available at \small\url{https://huggingface.co/datasets/weaverbirdllm/famma}, and the complete codebase---together with illustrative notebooks---is released at \small\url{https://github.com/famma-bench/bench-script}. 

\item {\bf Experimental setting/details}
    \item[] Question: Does the paper specify all the training and test details (e.g., data splits, hyperparameters, how they were chosen, type of optimizer, etc.) necessary to understand the results?
    \item[] Answer: \answerYes{} 
    \item[] Justification: the evaluation setting is clearly described at \cref{sec:exp_setup}. The training details of reasoning models can be found in \cref{app:training}.

\item {\bf Experiment statistical significance}
    \item[] Question: Does the paper report error bars suitably and correctly defined or other appropriate information about the statistical significance of the experiments?
    \item[] Answer: \answerYes{} 
    \item[] Justification: the evaluation is deterministic: all LLM inference is performed with temperature set to 0, eliminating sampling variance. Pass@1 accuracy is therefore constant across runs, therefore the error bars are ignored in reporting.

\item {\bf Experiments compute resources}
    \item[] Question: For each experiment, does the paper provide sufficient information on the computer resources (type of compute workers, memory, time of execution) needed to reproduce the experiments?
    \item[] Answer: \answerYes{} 
    \item[] Justification: the training resources of reasoning models is outlined in \cref{app:training}.
    
\item {\bf Code of ethics}
    \item[] Question: Does the research conducted in the paper conform, in every respect, with the NeurIPS Code of Ethics \url{https://neurips.cc/public/EthicsGuidelines}?
    \item[] Answer: \answerYes{} 
    \item[] Justification: we have read the guidelines and certify full compliance.

\item {\bf Broader impacts}
    \item[] Question: Does the paper discuss both potential positive societal impacts and negative societal impacts of the work performed?
    \item[] Answer: \answerYes{} 
    \item[] Justification: see \cref{app:impact}.
    
\item {\bf Safeguards}
    \item[] Question: Does the paper describe safeguards that have been put in place for responsible release of data or models that have a high risk for misuse (e.g., pretrained language models, image generators, or scraped datasets)?
    \item[] Answer: \answerYes{} 
    \item[] Justification: see \cref{app:safe}.

\item {\bf Licenses for existing assets}
    \item[] Question: Are the creators or original owners of assets (e.g., code, data, models), used in the paper, properly credited and are the license and terms of use explicitly mentioned and properly respected?
    \item[] Answer: \answerYes{} 
    \item[] Justification: we provide details of dataset curation in \cref{sec:benchmark}.

\item {\bf New assets}
    \item[] Question: Are new assets introduced in the paper well documented and is the documentation provided alongside the assets?
    \item[] Answer: \answerYes{} 
    \item[] Justification: the documentation is well prepared in the repository {\small \url{https://github.com/famma-bench/bench-script}} with markdown documents and notebooks.

\item {\bf Crowdsourcing and research with human subjects}
    \item[] Question: For crowdsourcing experiments and research with human subjects, does the paper include the full text of instructions given to participants and screenshots, if applicable, as well as details about compensation (if any)? 
    \item[] Answer: \answerYes{} 
    \item[] Justification: we provide details of dataset curation in \cref{sec:benchmark} and \cref{app:dataset}.

\item {\bf Institutional review board (IRB) approvals or equivalent for research with human subjects}
    \item[] Question: Does the paper describe potential risks incurred by study participants, whether such risks were disclosed to the subjects, and whether Institutional Review Board (IRB) approvals (or an equivalent approval/review based on the requirements of your country or institution) were obtained?
    \item[] Answer: \answerNA{} 
    \item[] Justification: see \cref{app:human}.

\item {\bf Declaration of LLM usage}
    \item[] Question: Does the paper describe the usage of LLMs if it is an important, original, or non-standard component of the core methods in this research? Note that if the LLM is used only for writing, editing, or formatting purposes and does not impact the core methodology, scientific rigorousness, or originality of the research, declaration is not required.
    \item[] Answer: \answerNA{} 
    \item[] Justification: LLMs are not employed in any component that affects the study’s core methodology, scientific rigor, or original contributions.

\end{enumerate}

\end{document}